\documentclass[journal]{IEEEtran}

\ifCLASSINFOpdf
\else
\fi

\usepackage{times}
\usepackage{epsfig}
\usepackage{graphicx}
\usepackage{amsmath}
\usepackage{amssymb}
\usepackage{mathrsfs}
\usepackage{multirow}
\usepackage{blkarray}
\usepackage{float}
\usepackage{array}
\usepackage{epstopdf}
\usepackage{subcaption}
\usepackage[T1]{fontenc}

\newcommand{\argmin}{\arg\!\min}

\begin{document}

\title{Image Decomposition Using a Robust Regression Approach}

\author{Shervin~Minaee,~\IEEEmembership{Student Member,~IEEE,}
        and~Yao~Wang,~\IEEEmembership{Fellow,~IEEE}
}

\maketitle

\begin{abstract}
This paper considers how to separate text and/or graphics from smooth background in screen content and mixed content images and proposes an algorithm to perform this segmentation task.
The proposed methods make use of the fact that the background in each block is usually smoothly varying and can be modeled well by a linear combination of a few smoothly varying basis functions, while the foreground text and graphics create sharp discontinuity. This algorithm separates the background and foreground pixels by trying to fit pixel values in the block into a smooth function using a robust regression method. 
The inlier pixels that can be well represented with the smooth model will be considered as background, while remaining outlier pixels will be considered foreground. 
We have also created a dataset of screen content images extracted from HEVC standard test sequences for screen content coding with their ground truth segmentation result which can be used for this task.
The proposed algorithm has been tested on the dataset mentioned above and is shown to have superior performance over other methods, such as the hierarchical k-means clustering algorithm, shape primitive extraction and coding, and the least absolute deviation fitting scheme for foreground segmentation.
\end{abstract}


\IEEEpeerreviewmaketitle

\section{Introduction}
Screen content images refer to images appearing on the display screens of electronic devices such as computers and smart phones \cite{SCC_tran}, \cite{SCC}.  These images have similar characteristics as mixed content documents (such as a magazine page). They often contain two layers, a pictorial smooth background and a foreground consisting of text and line graphics. The usual image compression algorithms such as JPEG2000 \cite{jpeg} and HEVC intra frame coding \cite{HEVC} may not result in a good compression rate for this kind of images because the foreground consists of sharp discontinuities. 
In these cases, segmenting the image into two layers and coding them separately may be more efficient.
The idea of segmenting an image for better compression was proposed for check image compression \cite{check}, in DjVu algorithm for scanned document compression \cite{djvu} and the mixed raster content representation \cite{raster}.
Foreground segmentation has also applications in medical image segmentation,  text extraction (which is essential for automatic character recognition and image understanding), biometrics recognition, and object of interest detection biomedical applications \cite{chromosome}-\cite{bio2}.

Screen content and mixed document images are hard to segment, because the foreground may be overlaid over a smoothly varying background that has a color range that overlaps with the color of the foreground. Also because of the use of sub-pixel rendering, the same text/line often has different colors. Even in the absence of sub-pixel rendering, pixels belonging to the same text/line often have somewhat different colors.

Different algorithms have been proposed in the past for foreground-background segmentation in still images such as hierarchical k-means clustering in DjVu \cite{djvu}, which applies the k-means clustering algorithm on a large block to obtain foreground and background colors and then uses them as the initial foreground and background colors for the smaller blocks in the next stages, shape primitive extraction and coding (SPEC) \cite{spec} which first classifies each block of size $16 \times 16$ into either pictorial block or text/graphics based on the number of colors and
then refines the segmentation result of pictorial blocks, by extracting shape primitives and then comparing the size and color of the shape primitives with some threshold, and least absolute deviation fitting \cite{LAD}.
There are also some recent algorithms based on sparse decomposition proposed for this task \cite{sp_sm}-\cite{SD_journal}.

Most of the previous works have difficulty for the regions where background and foreground color intensities overlap and some part of the background will be detected as foreground or the other way.

The proposed segmentation algorithm in this work uses robust regression \cite{robustregression} techniques to overcome the problems of previous segmentation algorithms, which to the best of our knowledge has not been investigated previously. We model the background part of the image with a smooth function, by fitting a smooth model to the intensities of the majority of the pixels in each block. Any pixel whose intensity could be predicted well using the derived model would be considered as background and otherwise it would be considered as foreground. 
RANSAC algorithm is used here which is a powerful and simple robust regression technique.
To boost the speed of the algorithm, we also proposed some pre-processing steps which first check if a block can be segmented using some simpler approaches and it goes to RANSAC only if the block cannot be segmented using those approaches.
The proposed algorithm has various applications including, text extraction, segmentation based video coding and medical image segmentation.

The structure of the rest of this paper is as follows: Section II presents the proposed robust regression technique for foreground-background segmentation. The final segmentation algorithm that includes both the core robust regression algorithm as well as preprocessing steps is discussed in Section III. Section IV provides the experimental results for these algorithms. And finally the paper is concluded in Section V.

\section{Background Modeling and Robust Regression}
We assume that if an image block only consists of background, it should be well represented with a few smooth basis functions. By well representation we mean that the approximated value at a pixel with the smooth functions should have an error less than a desired threshold at every pixel. But if an image block has some foreground pixels overlaid on top of a smooth background, and these foreground pixels occupy a relatively small percentage of the block, then the fitted smooth function will not represent these foreground pixels well.

To be more specific, we divide each image into non-overlapping blocks of size $N\times N$, and represent each image block, denoted by $F(x,y)$, with a smooth model $B(x,y;\alpha_1,...,\alpha_K)$, where $x$ and $y$ denote the horizontal and vertical axes and $\alpha_1,...,\alpha_K$ denote the parameters of this smooth model. 

Two questions should be addressed here, the first one is how to find a suitable smooth model $B(x,y;\alpha_1,...,\alpha_K)$, and the second one is how to find the optimal value of parameters of our model such that they are not affected by foreground pixels, especially if we have many foreground pixels.
We order all the possible basis functions in the conventional zig-zag order in the $(u,v)$ plane, and choose the first $K$ basis functions.

For the first question, following the work in \cite{LAD} we use a linear combination of some basis functions $P_k(x,y)$, so that the model can be represented as $\sum_{k=1}^K \alpha_k P_k(x,y)$. 
Then we used the Karhunen-Loeve transform \cite{KLT} on a set of training images that only consist of smooth background to derive the optimum set of bases. The derived bases turned out to be very similar to 2D DCT basis functions. Because of that we decided to use a linear combination of a set of $K$ 2D DCT bases as our smooth model.
The 2-D DCT function is defined as:
\begin{equation*}
P_{u,v}(x,y)= \beta_u \beta_v cos((2x+1)\pi u/2N) cos((2y+1)\pi v/2N) 
\end{equation*}
where $u$ and $v$ denote the frequency of the basis and $\beta_u$ and $\beta_v$ are normalization factors.
It is good to note that algorithms based on supervised dictionary learning and subspace learning are also useful for deriving the smooth representation of background component \cite{rahmani1}-\cite{gholami2}.

The second question is a chicken-and-egg problem: To find the model parameters we need to know which pixels belong to the background; and to know which pixels belong to background we need to know what are the model parameters. 
One solution to find the optimal model parameters, $\alpha_k$'s, is to define some cost function, which measures the goodness of fit between the original pixel intensities and the ones predicted by the smooth model, and then minimize the cost function. 
One plausible cost function can be the $\ell_p$-norm of the fitting error ($p$ can be 0, 1, or 2), so that the solution can be written as:
\begin{gather*}
\{\alpha_1^*,...,\alpha_K^*\}= \argmin_{\alpha_1,...,\alpha_K}  \sum_{x,y} |F(x,y)- \sum_{k=1}^K \alpha_k P_k(x,y)|^p
\end{gather*}
Let $\boldsymbol{f}$, $\boldsymbol{\alpha}$ and $\textbf{P}$ denote the 1D version of $F$, the vector of all parameters and a matrix of size $N^2\times K$ in which the k-th column corresponds to the vectorized version of $P_k(x,y)$ respectively.
Then the above problem can be formulated as $\boldsymbol{\alpha}^*= \argmin_{\boldsymbol{\alpha}} \| \boldsymbol{f}-\textbf{P}\boldsymbol{\alpha} \|_p$.\\
Now if we use the $\ell_2$-norm (i.e. $p=2$) for the cost function we simply get the least squares fitting problem and, which has a closed-form solution as below:
\begin{gather}
\boldsymbol{\alpha}= (\textbf{P}^T \textbf{P})^{-1}\textbf{P}^T \boldsymbol{f}
\end{gather}
But the least square fitting suffers from the fact that the model parameters, $\boldsymbol{\alpha}$, can be adversely affected by foreground pixels.

Here we propose an alternative method based on robust regression, which tries to minimize the the number of outliers and fitting the model only to inliers. 
The notion of robustness is greatly used in computer vision, for fundamental matrix and object recognition \cite{fund1}-\cite{rahimi}.
RANSAC algorithm is used in this work, which is more robust to outliers and the resulting model is less affected by them. This algorithm is explained below.

\subsection{RANSAC Based Segmentation}
RANSAC \cite{RANSAC} is a popular robust regression algorithm which is designed to find the right model for a set of data even in the presence of outliers. RANSAC is an iterative approach that performs the parameter estimation by minimizing the number of outliers (which can be thought as minimizing the $\ell_0$-norm). We can think of foreground pixels as outliers for the smooth model in our segmentation algorithm.
RANSAC repeats two iterative procedures to find a model for a set of data.
In the first step, it takes a subset of the data and derives the parameters of the model only using that subset. 
In the second step, it tests the model derived from the first step against the entire dataset to see how many samples can be modeled consistently. 
A sample will be considered as an outlier if it has a fitting error larger than a threshold that defines the maximum allowed deviation. 
RANSAC repeats the procedure a fixed number of times and at the end, it chooses the model with the largest consensus set (the set of inliers) as the optimum model. 
The proposed RANSAC algorithm for foreground/background segmentation of a block of size $N \times N$ is as follows:
\begin{enumerate}
\item Select a subset of $K$ randomly chosen pixels. Let us denote this subset by $S=\{(x_l,y_l), \ l=1,2,\ldots,K\}$.
\item Fit the model $\sum_{k=1}^K \alpha_k P_k(x,y) $ to the pixels $(x_l,y_l) \in S$ and find the $\alpha_k$'s. This is done by solving the set of $K$ linear equations $\sum_k \alpha_k P_k(x_l,y_l) = F(x_l,y_l), \ l=1,2,\ldots,K$.
Here $F(x,y)$ denotes the luminance value at pixel $(x,y)$.
\item Test all $N^2$ pixels $F(x,y)$ in the block against the fitted model. Those pixels that can be predicted with an error less than $\epsilon_{in}$ will be considered as the inliers.
\item Save the consensus set of the current iteration if it has a larger size than the previous one.
\item Repeat this procedure up to $M_{\rm iter}$ times, or when the largest concensus set found occupies over a certain percentage of the entire dataset, denoted by $\epsilon_2$.
\end{enumerate}
After this procedure is finished, the pixels in the largest consensus set will be considered as inliers or equivalently background.


\section{Overall Segmentation Algorithms}
We propose a segmentation algorithm that mainly depends on RANSAC but it first checks if a block can be segmented using some simpler approaches and it goes to RANSAC only if the block cannot be segmented using those approaches. These simple cases belong to one of these groups: nearly constant blocks, smoothly varying background and text/graphic overlaid on constant background.
 
Nearly constant blocks are those in which  all pixels have similar intensities. If the standard deviation of a block is less than some threshold we declare that block as nearly constant.
Smoothly varying background is a block in which the intensity variation over all pixels can be modeled well by a smooth function. Therefore we try to fit $K$ DCT basis to all pixels using least square fitting. If all pixels of that block can be represented with an error less than a predefined threshold, $\epsilon_{in}$, we declare it as smooth background.
The image blocks belonging to the text/graphic overlaid on constant background usually have zero variance (or very small variances) inside each connected component. These images usually have a limited number of different colors in each block (usually less than 10) and the intensities in different parts are very different. We calculate the percentage of each different color in that block and the one with the highest percentage will be chosen as background and the other ones as foreground. 
When a block does not satisfy any of the above conditions, RANSAC will be applied to separate the background and the foreground.

The overall segmentation algorithm for each blocks of size $N\times N$ is summarized as follows (Note that we only apply the algorithm to the gray scale component of a color image):
\begin{enumerate}
\item If the standard deviation of pixels' intensities is less than $\epsilon_1$, then declare the entire block as background.
If not, go to the next step;
\item Perform least square fitting using all pixels. If all pixels can be predicted with an error less than $\epsilon_{in}$, declare the entire block as background. If not, go to the next step;
\item If the number of different colors is less than $T_1$ and the intensity range is above $R$, declare the block as text/graphics over a constant background and use the color that has the highest percentage of pixels as the background color. If not, go to the next step;
\item Use RANSAC to segment background and foreground. Those pixels with fitting error less than $\epsilon_{in}$ will be considered as background.
\end{enumerate}

\section{Experimental Results}
To perform experimental studies we have generated an annotated dataset consisting of 332 image blocks of size $64\times 64$, extracted from HEVC test sequences for screen content coding \cite{SCC_data}. We have also manually extracted the ground truth foregrounds for these images. This dataset is publicly available at \cite{our_dataset}.

In our experiment, the block size is chosen to be $N$=64. The number of DCT basis functions, $K$, is set to be 10 based on prior experiments on a separate validation dataset \cite{LAD}. The inlier maximum allowed distortion is chosen as $\epsilon_{in}=10$. 
The maximum number of iteration in RANSAC algorithm is chosen to be $M_{\rm iter}=200$.
The thresholds used for preprocessing (steps 1-3) should be chosen conservatively to avoid segmentation errors. In our simulations, we have chosen them as $\epsilon_1=3$, $T_1=10$, $R=50$ and $\epsilon_2=0.95$, which achieved a good trade off between computation speed and segmentation accuracy.

To illustrate the smoothness of the background layer and its suitability for being coded with transform-based coding, the filled background layer of a sample image is presented in Figure 1. The background holes (those pixels that belong to foreground layers) are filled by the predicted value using the smooth model, which is obtained using the least squares fitting to the detected background pixels. As we can see the background layer is very smooth and does not have any sharp edges.
\begin{figure}[h]
\begin{center}
\hspace{-0.1cm}
    \includegraphics [scale=0.21] {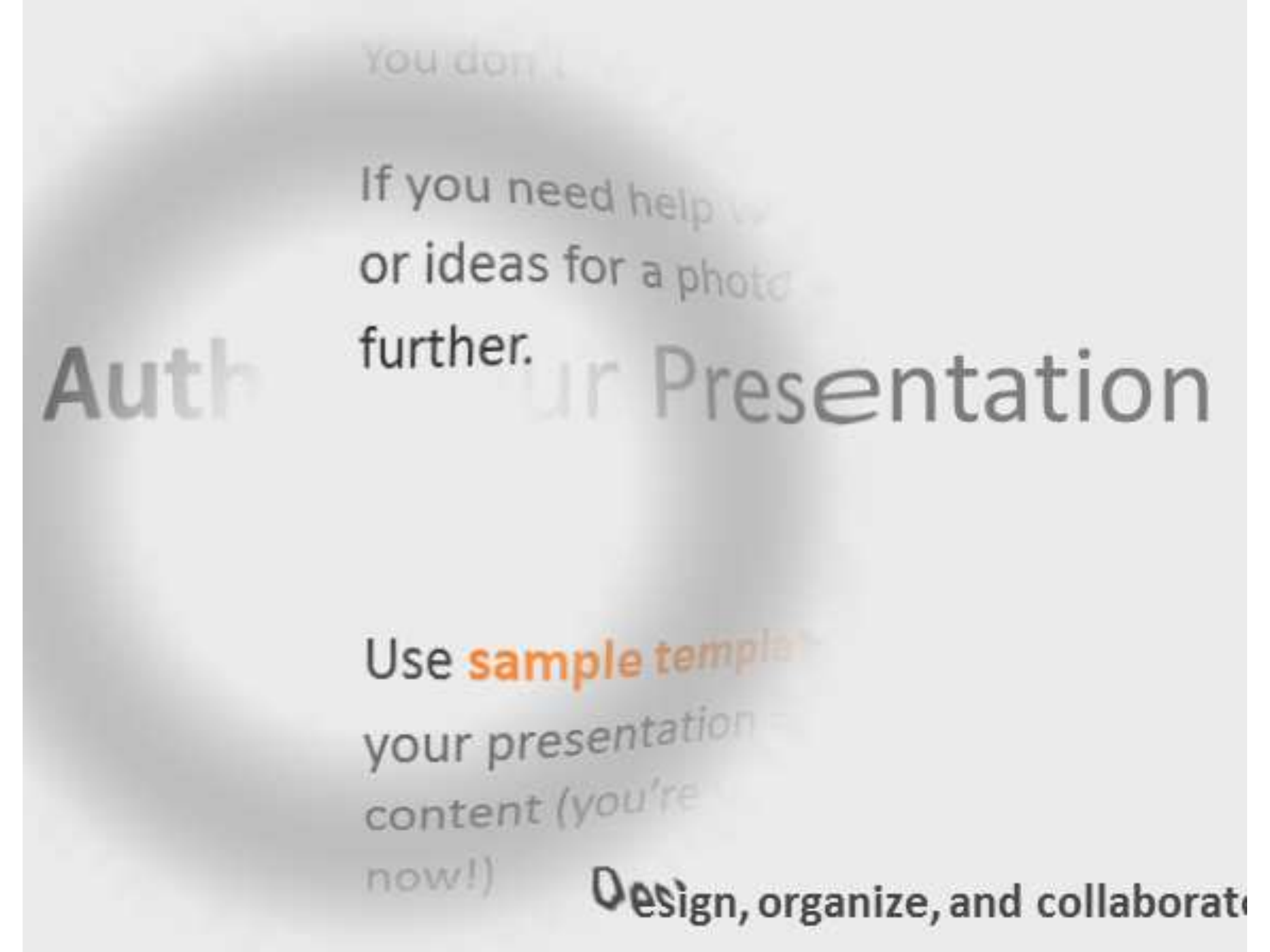}
    \vspace{0.3cm}
\hspace{-0.18cm}	\includegraphics [scale=0.21] {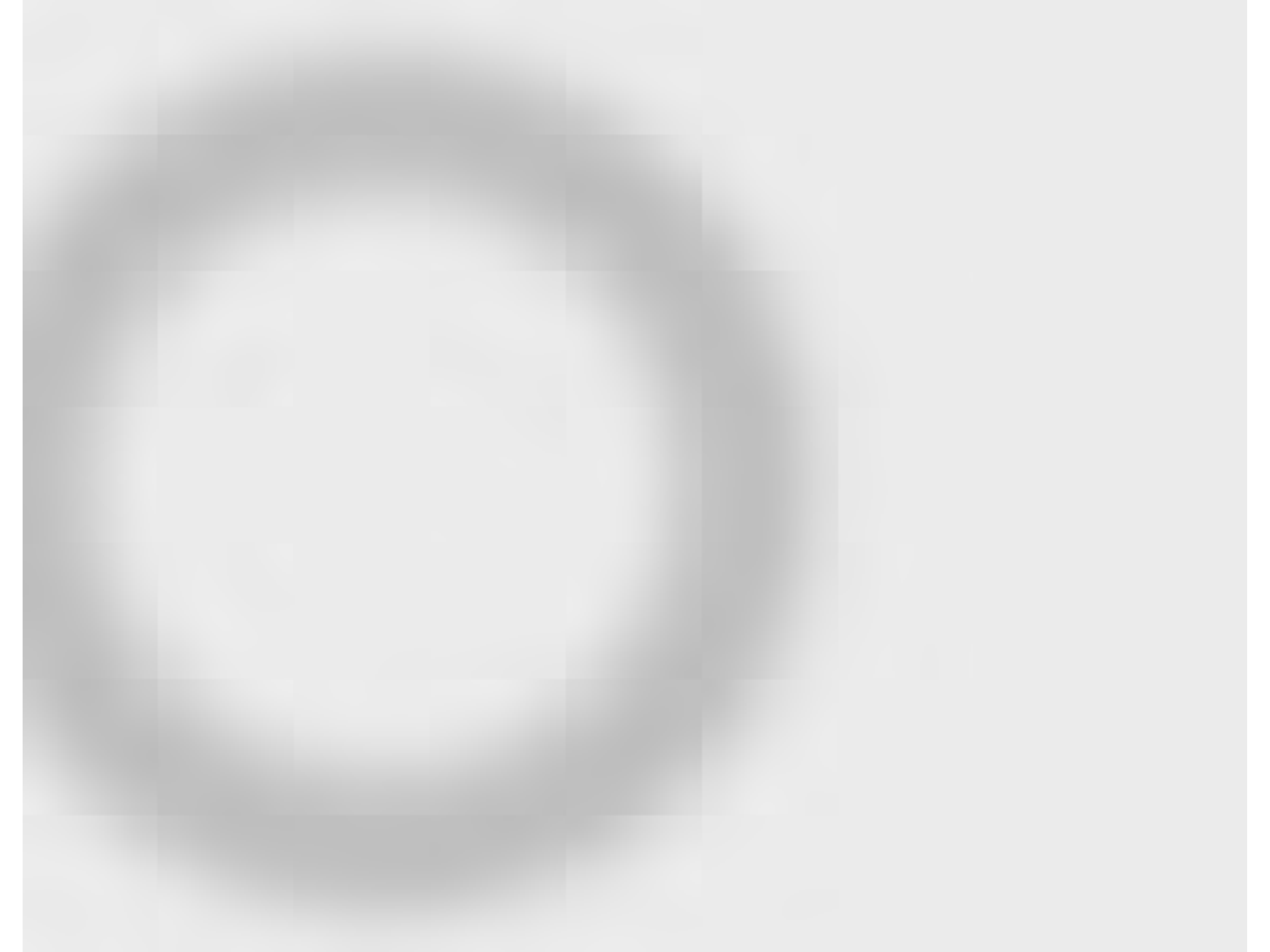}
\end{center}
  \caption{Left: the original image. Right: the reconstructed background of that.  The segmented foreground for this image can be found in Fig 2. }
\end{figure}

We have compared the proposed approach with three previous algorithms; least absolute deviation fitting, hierarchical k-means clustering and SPEC.
We have also provided a comparison with least square fitting algorithm result, so that the reader can see the benefit of minimizing the $\ell_0$ norm over the $\ell_2$ norm for model fitting.

To provide a numerical comparison between the proposed scheme and previous approaches, we have calculated the average precision, recall, and F1 score (also known as F-measure) achieved by different segmentation algorithms over this dataset. These results are presented in Table 1. 
The precision and recall are defined as in Eq. (2), where $\text{TP}, \text{FP}$ and $\text{FN}$ denote true positive, false positive and false negative respectively. In our evaluation, we treat a foreground pixel as positive. A pixel that is correctly identified as foreground (compared to the manual segmentation) is considered true positive. The same holds for false negative and false positive. 
\begin{gather}
 \text{Precision}= \frac{\text{TP}}{\text{TP+FP}} \ , 
\ \ \ \ \text{Recall}= \frac{\text{TP}}{\text{TP+FN}} 
\end{gather}
The balanced F1 score is defined as the harmonic mean of precision and recall.
\begin{gather*}
\text{F1}= 2 \ \frac{\text{precision} \times \text{recall}}{\text{precision+recall}}
\end{gather*}

As it can be seen, the proposed scheme  achieves  higher precision and recall and F1 score than other algorithms. 
\begin{table}[h]
\centering
  \caption{Accuracy comparison of different algorithms}
  \centering
\begin{tabular}{|m{4.1cm}|m{1.2cm}|m{1cm}|m{1cm}|}
\hline
Segmentation Algorithm  &  \  \ Precision & \ \  Recall &  F1 score\\
\hline
SPEC \cite{spec} & \ \ \ 50\% & \ \ \  64\% & \ \ \ 56\% \\
\hline
 Hierarchical Clustering \cite{djvu} & \ \ \ 64\% & \ \ \ 69\% & \ \ \ 66\% \\
\hline
 Least square fitting & \ \ \ 79\% & \ \ \ 60\% & \ \  \ 68\% \\ 
\hline
 Least Absolute Deviation \cite{LAD} & \ \ \  91.4\% & \ \ \  87\% & \ \ \  89.1\% \\
\hline
 Sparse-smooth decomposition \cite{sp_sm} & \ \ \ 64\% & \ \ \ 95\% & \ \  \ 76.4\% \\ 
\hline
 RANSAC based segmentation & \ \ \ 91.5\%  & \ \ \ 90\%  & \ \ \  90.7\%\\
\hline
\end{tabular}
\label{TblComp}
\end{table}

The results for 5 test images (each consisting of multiple 64x64 blocks) are shown in Fig. 2. 

\begin{figure*}[ht]
        \centering
        \vspace{-0.5cm}
        \begin{subfigure}[b]{0.18\textwidth}
                \includegraphics[width=\textwidth]{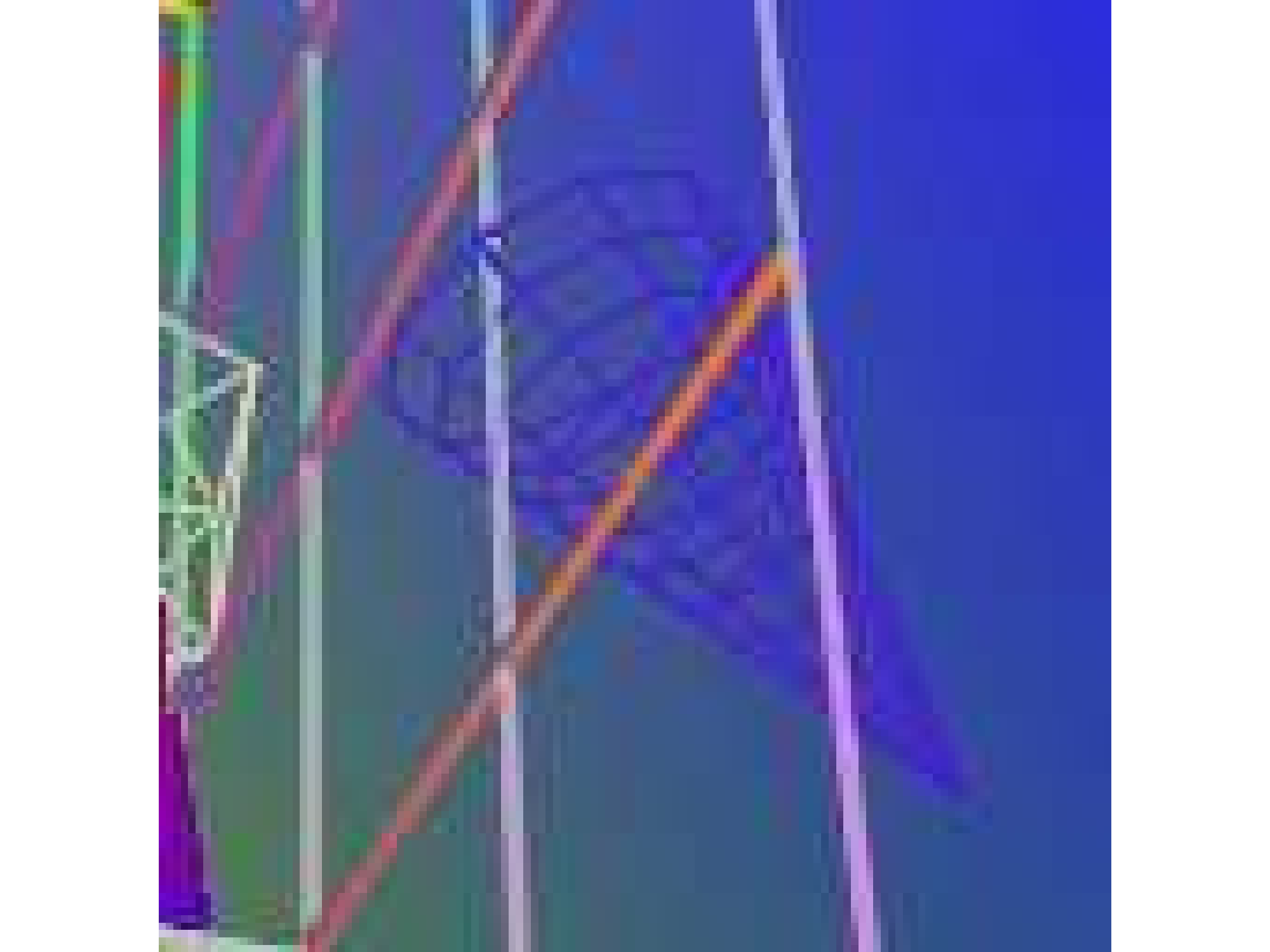}
                                \vspace{-0.5cm}
          \hspace{-2.5cm}    
        \end{subfigure}%
        ~ 
        \begin{subfigure}[b]{0.18\textwidth}
                \includegraphics[width=\textwidth]{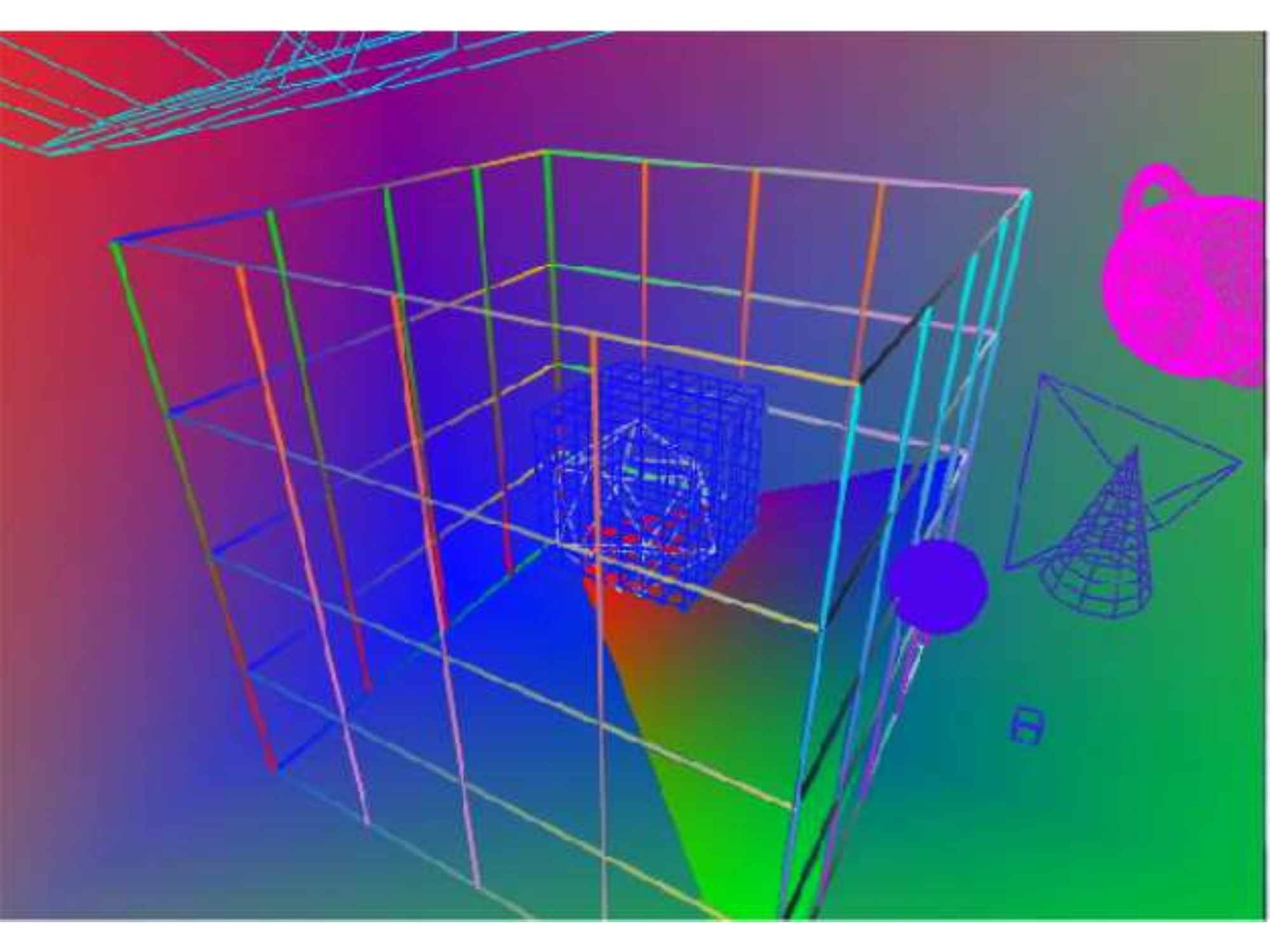}
                \vspace{-0.5cm}
            \hspace{-3cm} 
        \end{subfigure}%
        ~ 
        \begin{subfigure}[b]{0.18\textwidth}
                \includegraphics[width=\textwidth]{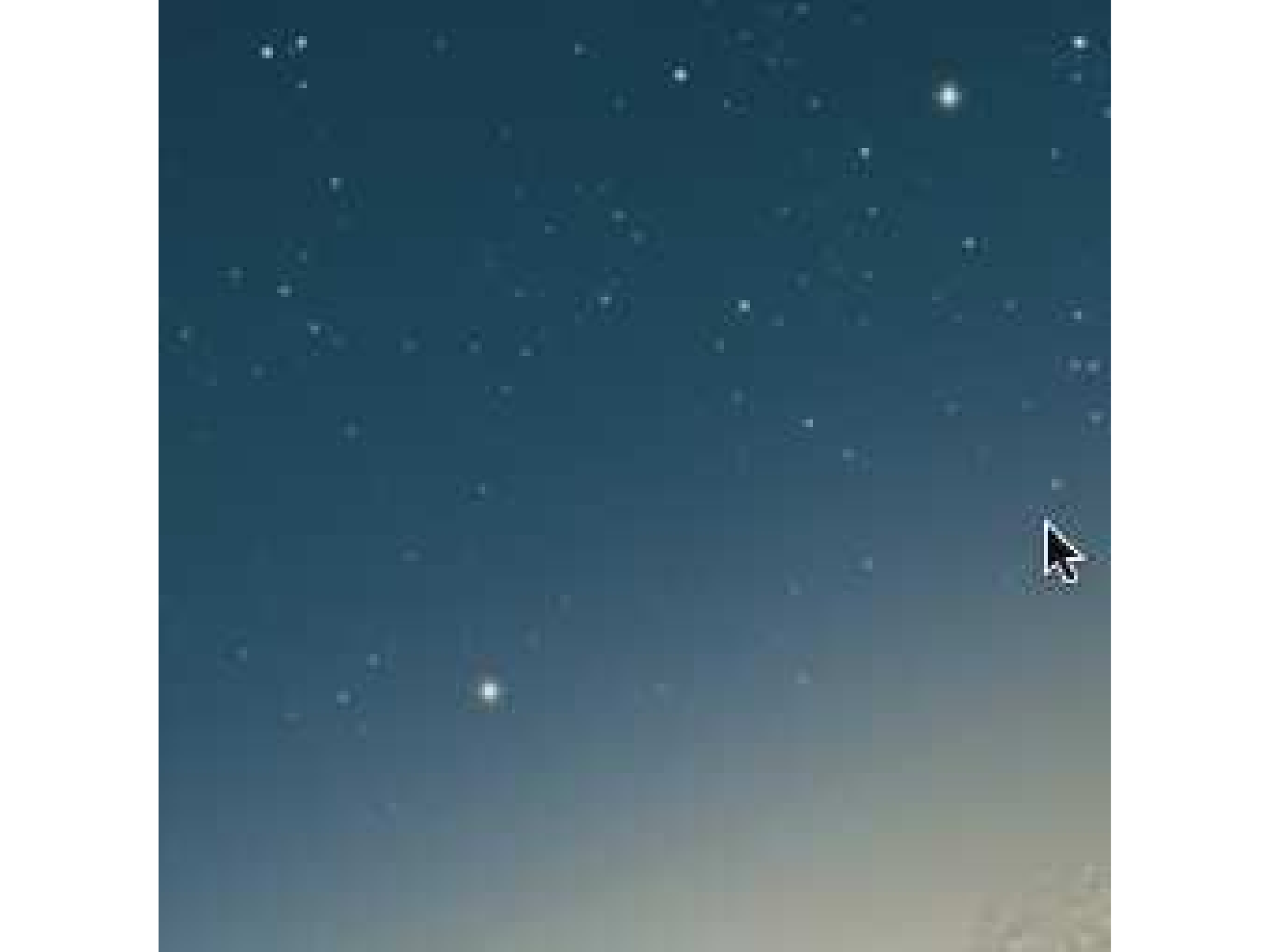}
                \vspace{-0.45cm}
            \hspace{-3cm} 
        \end{subfigure}%
        \begin{subfigure}[b]{0.18\textwidth}
			~ 
                \includegraphics[width=\textwidth]{2Original_Image-eps-converted-to.pdf}
                \vspace{-0.45cm}
            \hspace{-5cm} 
        \end{subfigure}%
        \begin{subfigure}[b]{0.18\textwidth}
                \includegraphics[width=\textwidth]{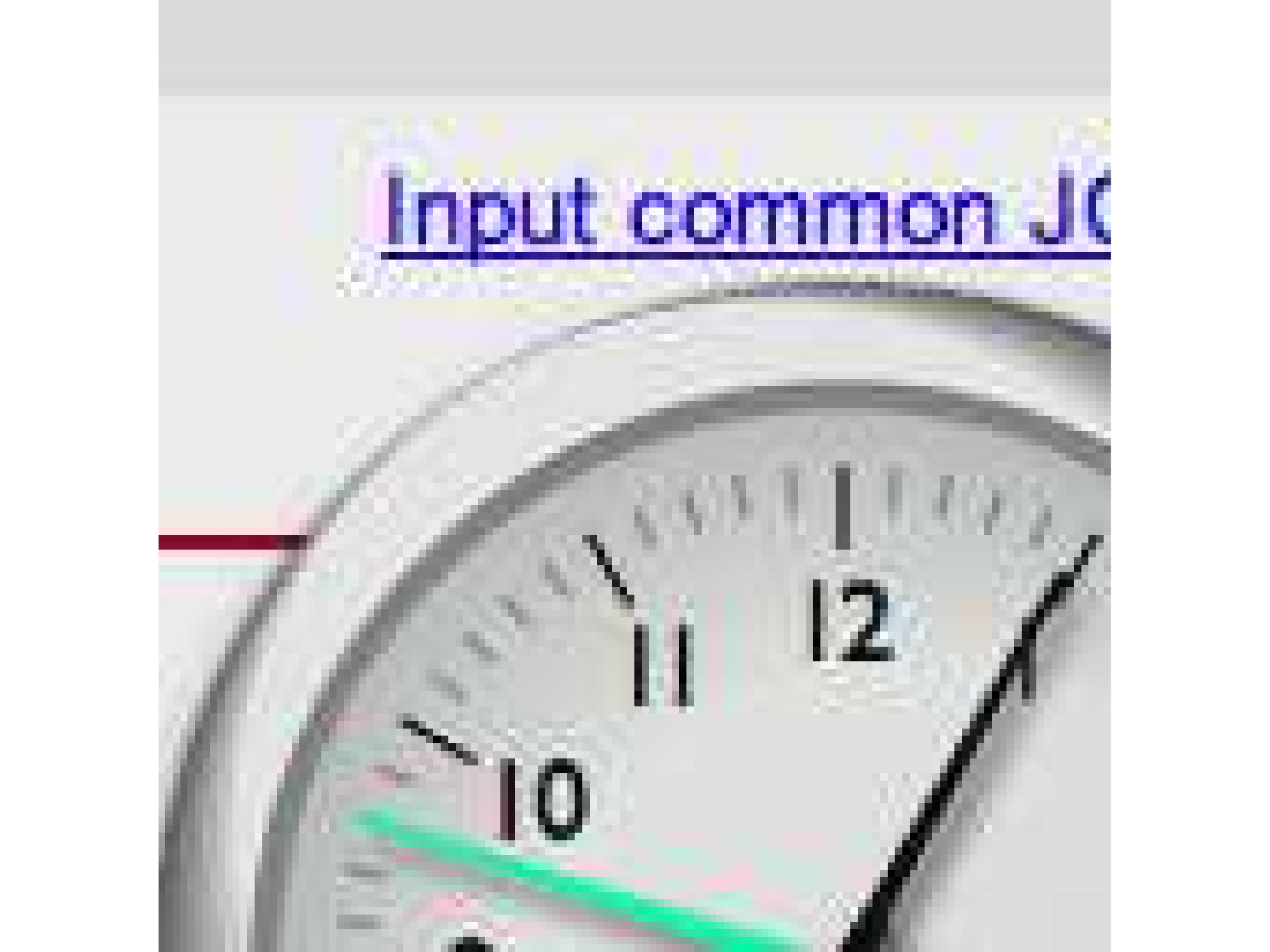}
                 \vspace{-0.45cm}
              \hspace{-4.8cm}
        \end{subfigure}
         \\[1ex]
        \begin{subfigure}[b]{0.18\textwidth}
                \includegraphics[width=\textwidth]{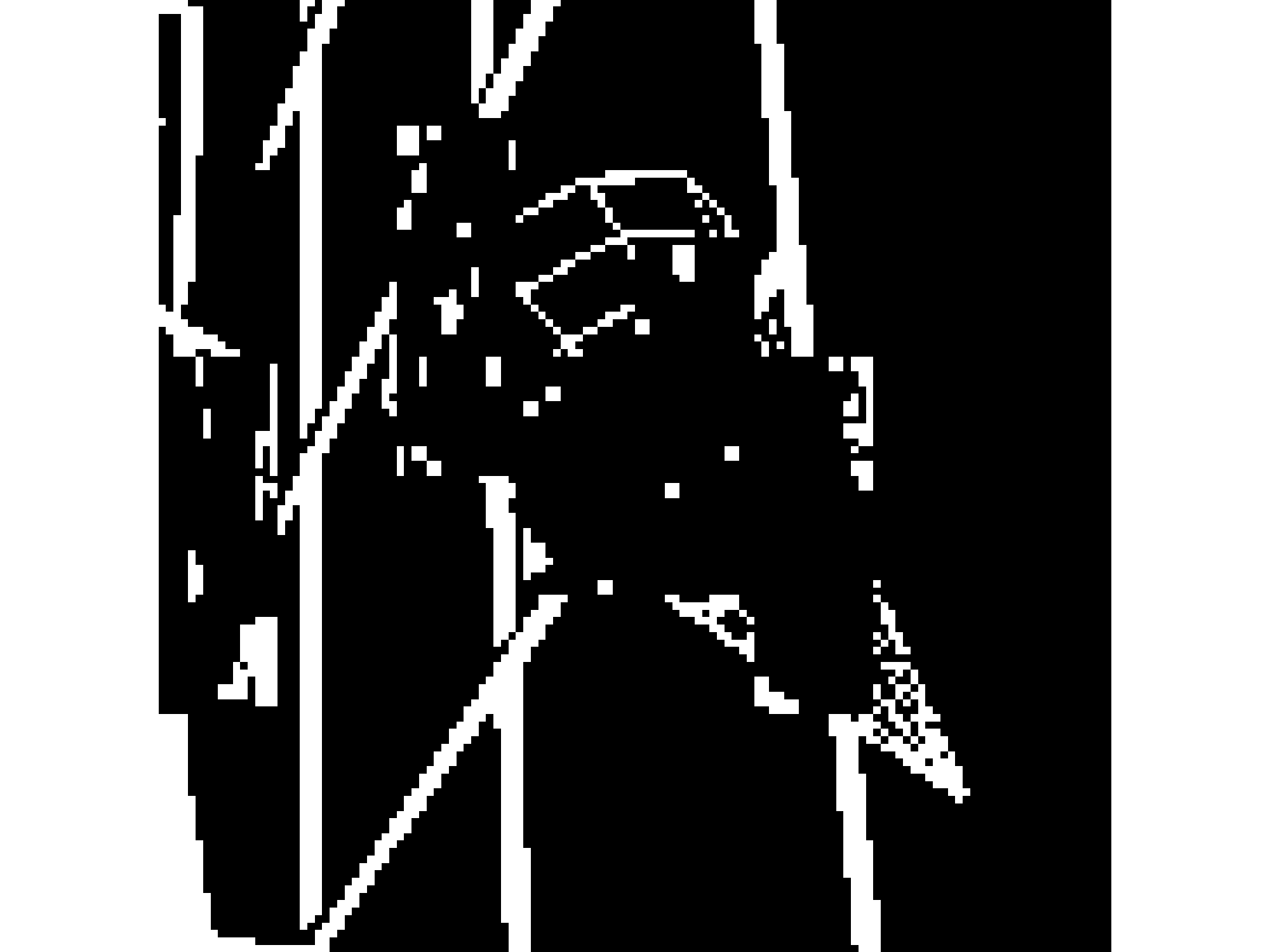}
                                \vspace{-0.5cm}
          \hspace{-2.5cm}    
        \end{subfigure}%
        ~ 
        \begin{subfigure}[b]{0.18\textwidth}
                \includegraphics[width=\textwidth]{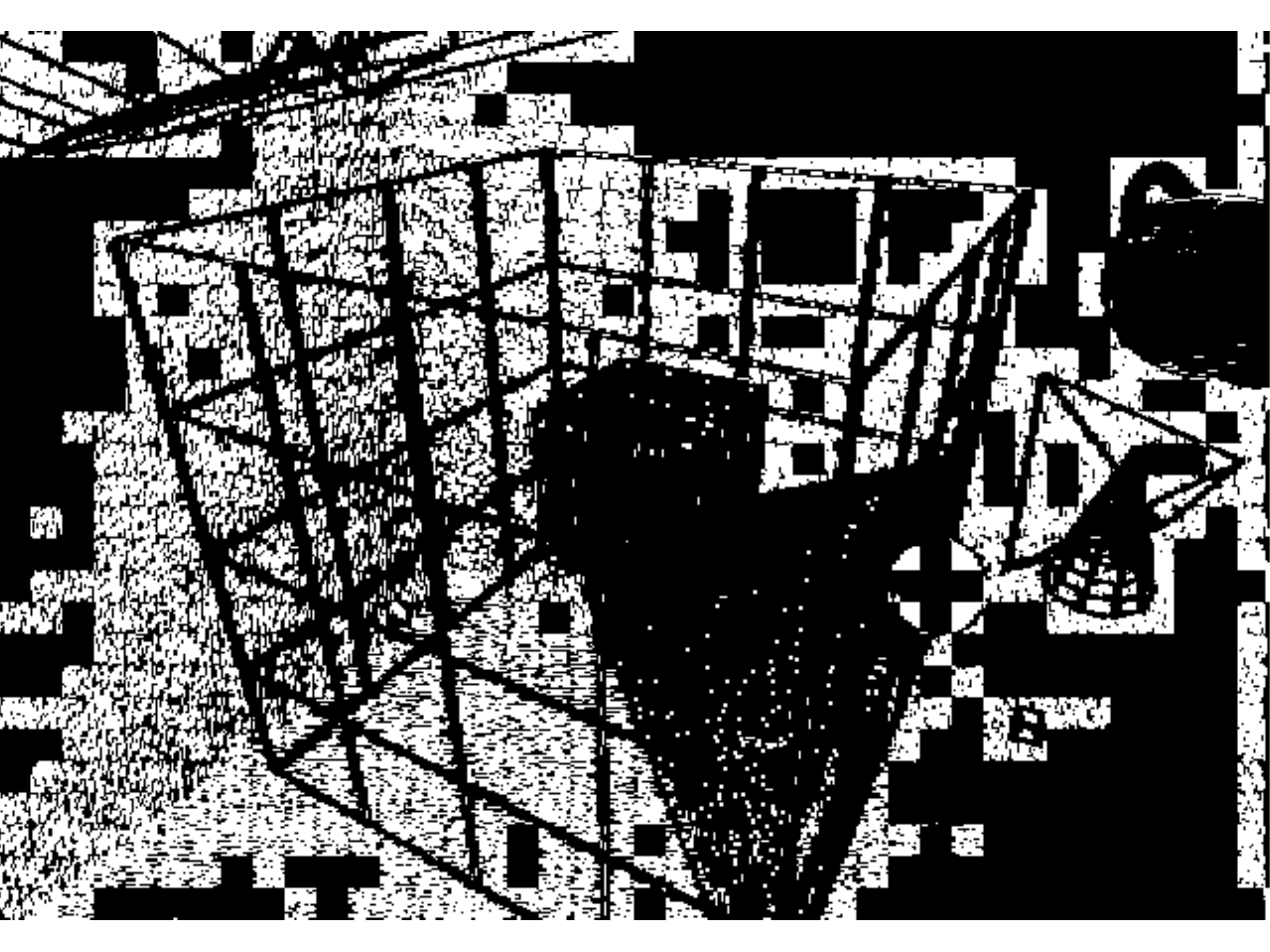}
                \vspace{-0.5cm}
            \hspace{-3cm} 
        \end{subfigure}%
        ~ 
        \begin{subfigure}[b]{0.18\textwidth}
                \includegraphics[width=\textwidth]{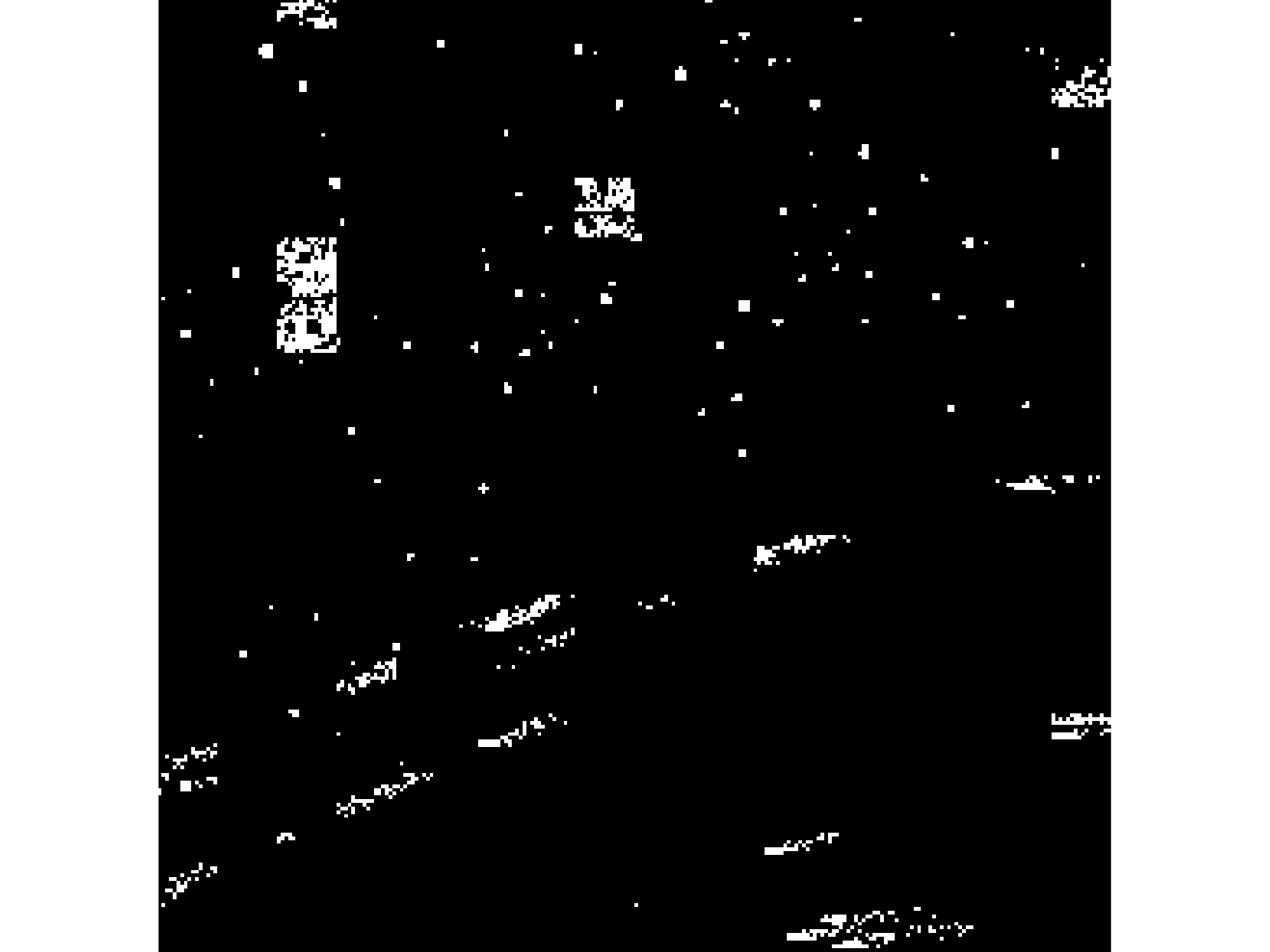}
                \vspace{-0.45cm}
            \hspace{-3cm} 
        \end{subfigure}%
        \begin{subfigure}[b]{0.18\textwidth}
			~ 
                \includegraphics[width=\textwidth]{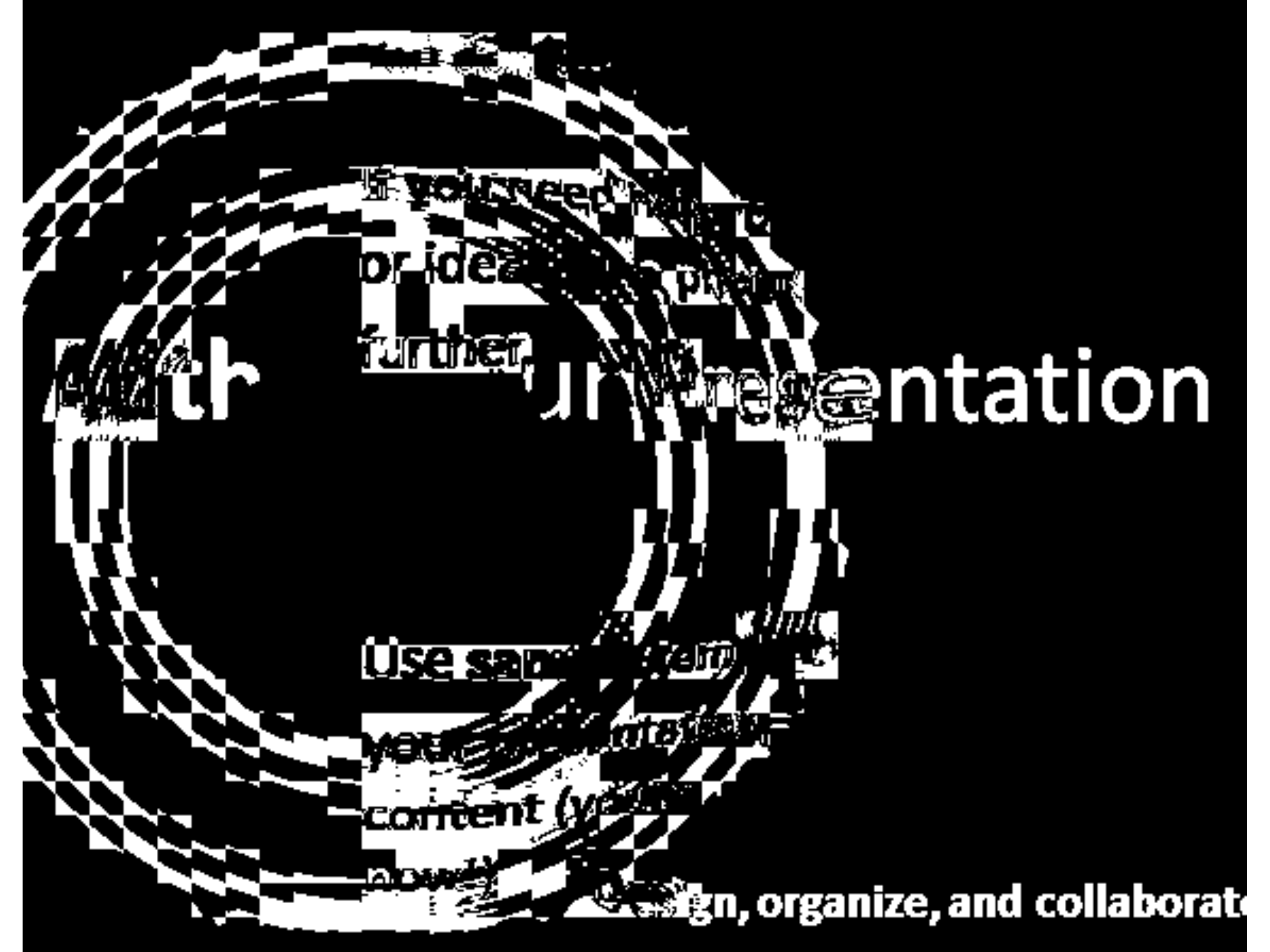}
                \vspace{-0.45cm}
            \hspace{-3cm} 
        \end{subfigure}%
        \begin{subfigure}[b]{0.18\textwidth}
                \includegraphics[width=\textwidth]{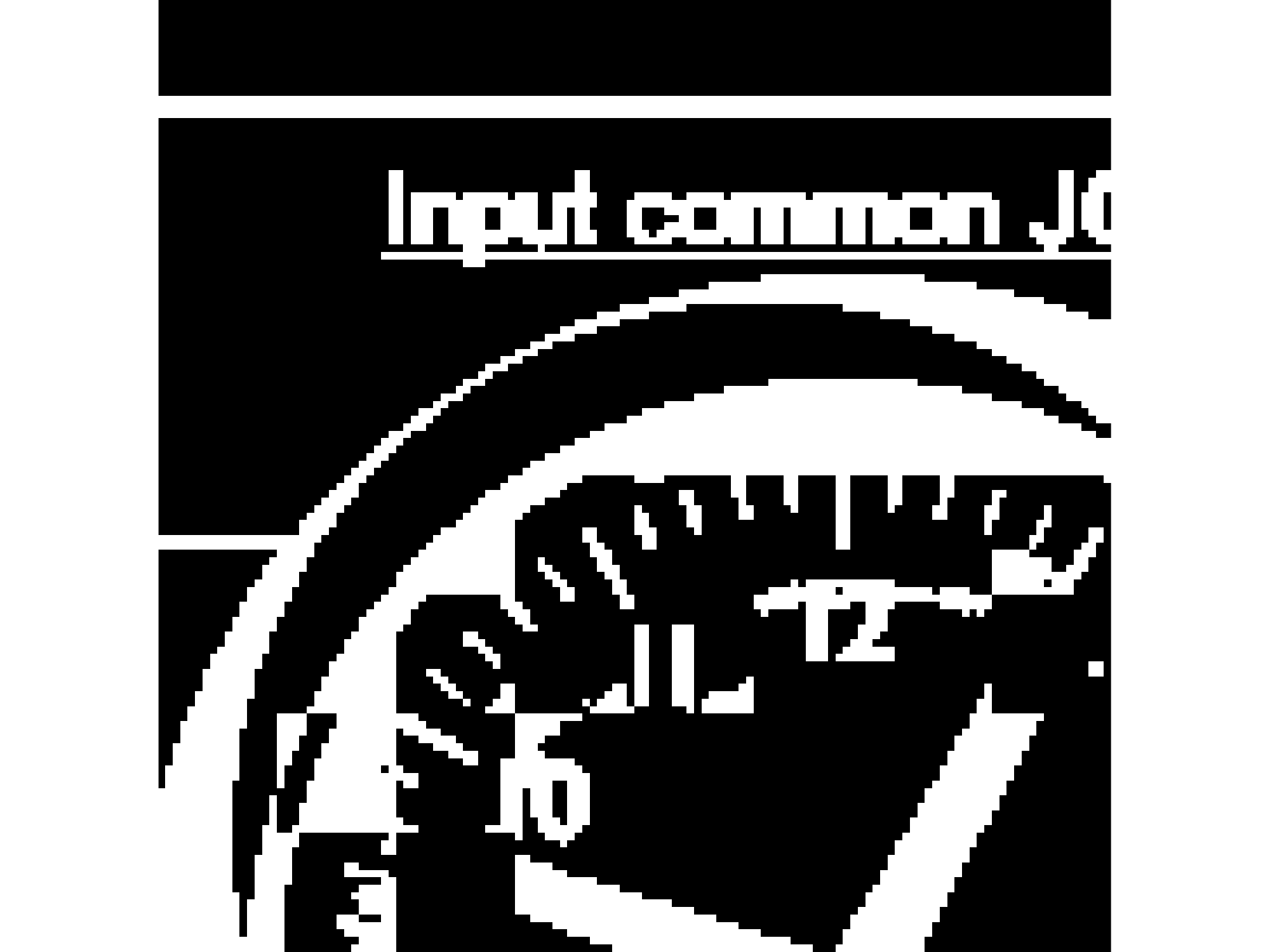}
                 \vspace{-0.45cm}
              \hspace{-4.8cm}
        \end{subfigure} \\[1ex]
        \begin{subfigure}[b]{0.18\textwidth}
                \includegraphics[width=\textwidth]{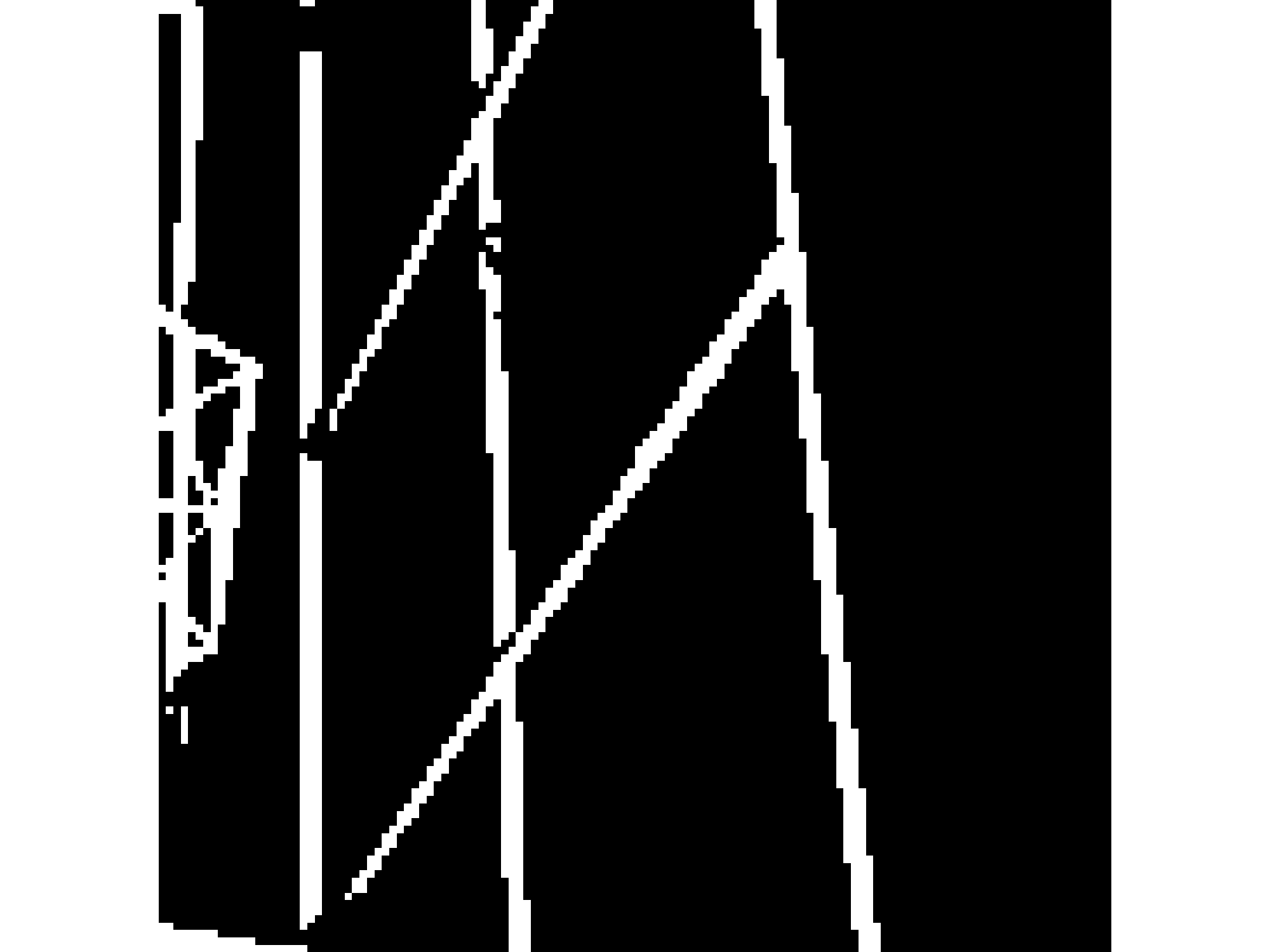}
                                \vspace{-0.5cm}
          \hspace{-2.5cm}    
        \end{subfigure}%
        ~ 
        \begin{subfigure}[b]{0.18\textwidth}
                \includegraphics[width=\textwidth]{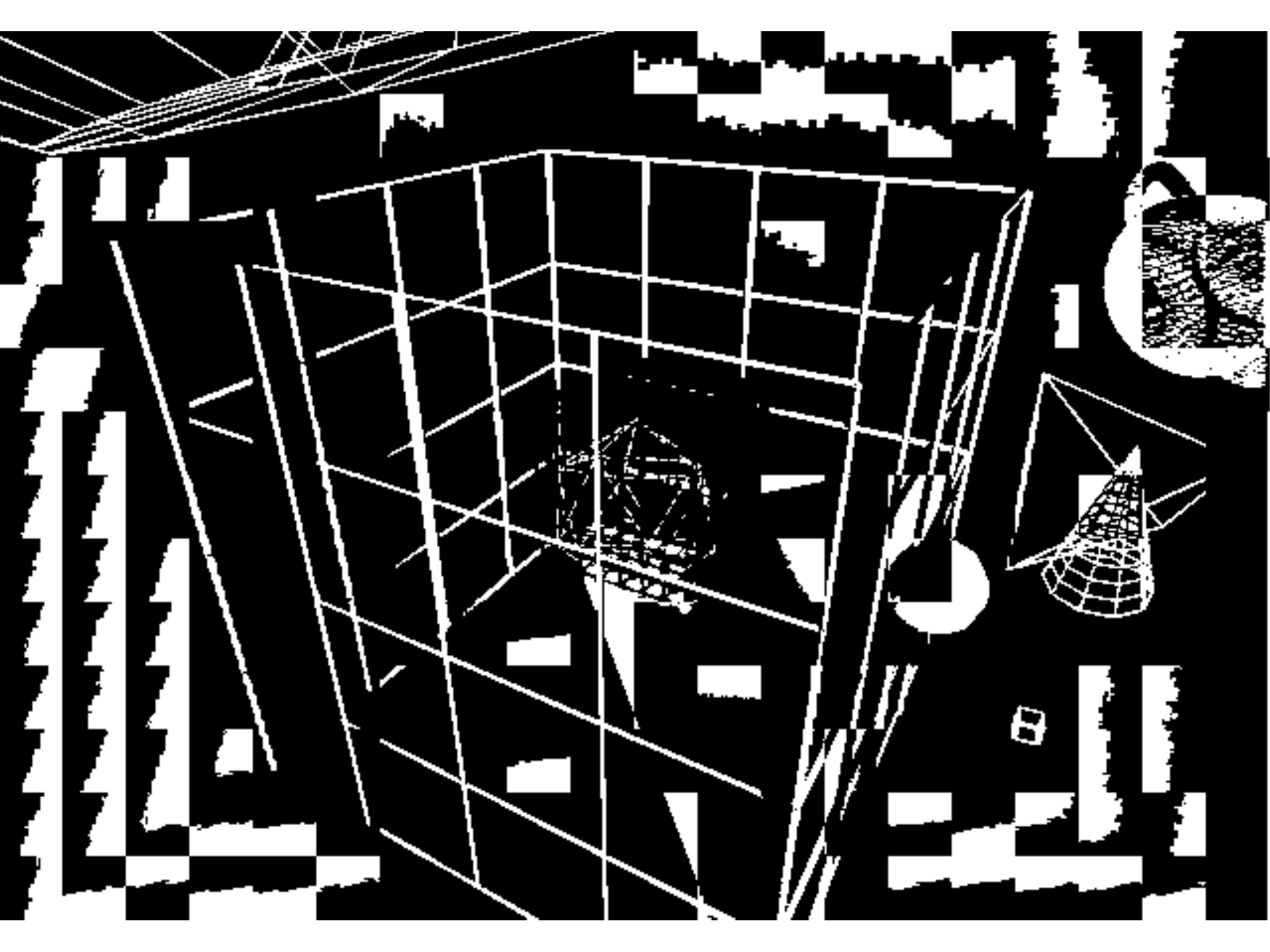}
                \vspace{-0.5cm}
            \hspace{-3cm} 
        \end{subfigure}%
        ~ 
        \begin{subfigure}[b]{0.18\textwidth}
                \includegraphics[width=\textwidth]{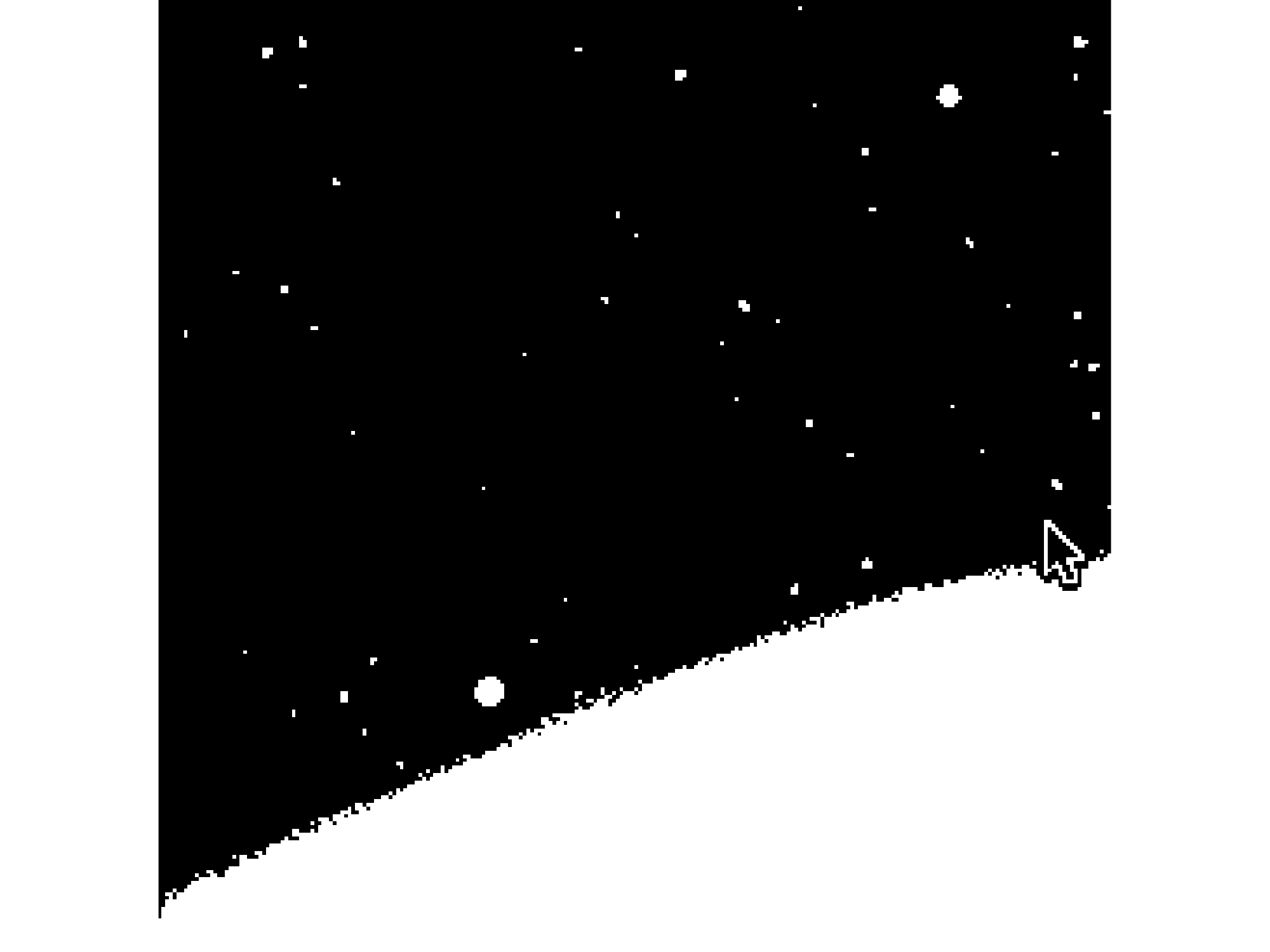}
                \vspace{-0.45cm}
            \hspace{-3cm} 
        \end{subfigure}%
        \begin{subfigure}[b]{0.18\textwidth}
			~ 
                \includegraphics[width=\textwidth]{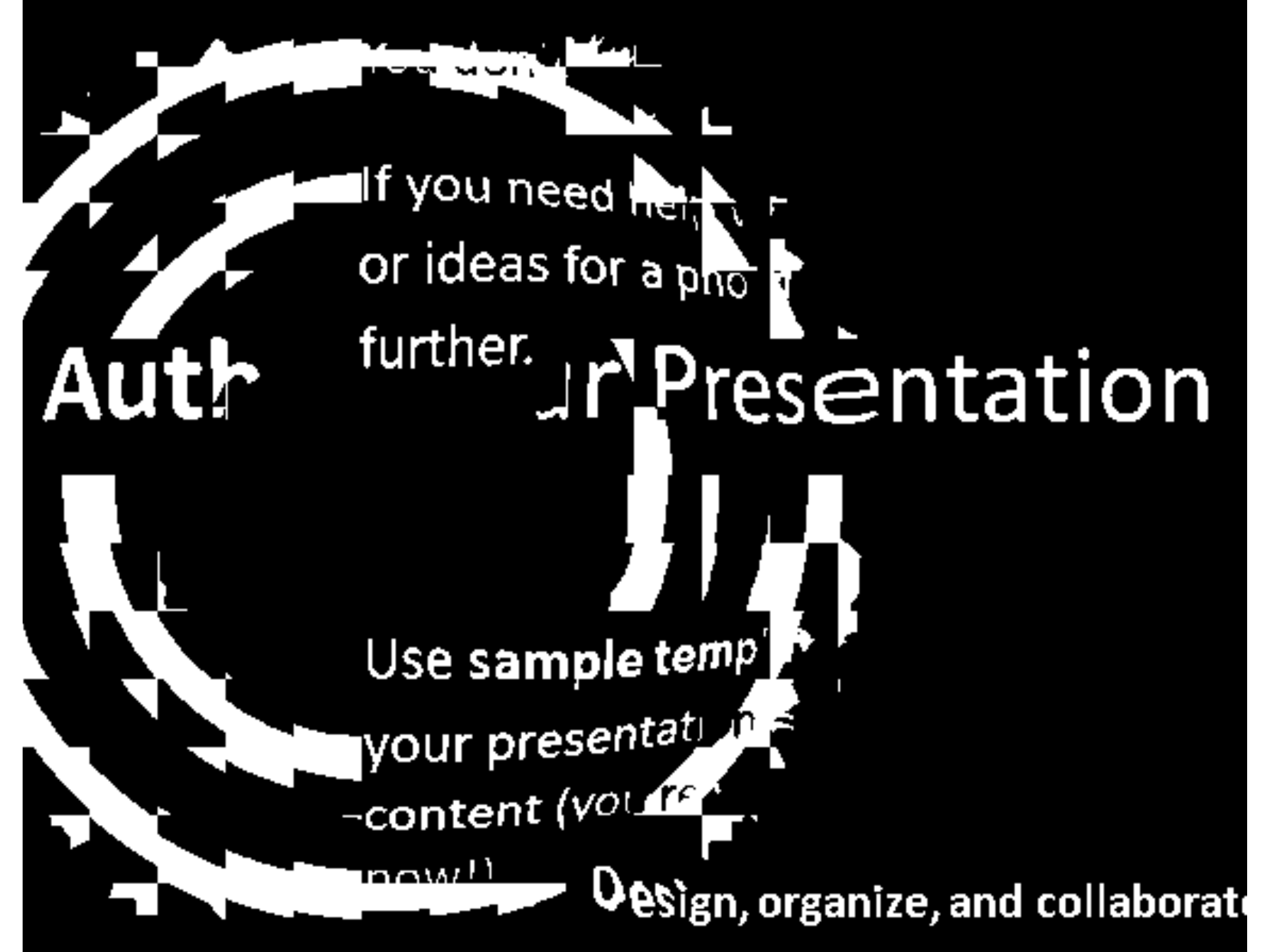}
                \vspace{-0.45cm}
            \hspace{-3cm} 
        \end{subfigure}%
        \begin{subfigure}[b]{0.18\textwidth}
                \includegraphics[width=\textwidth]{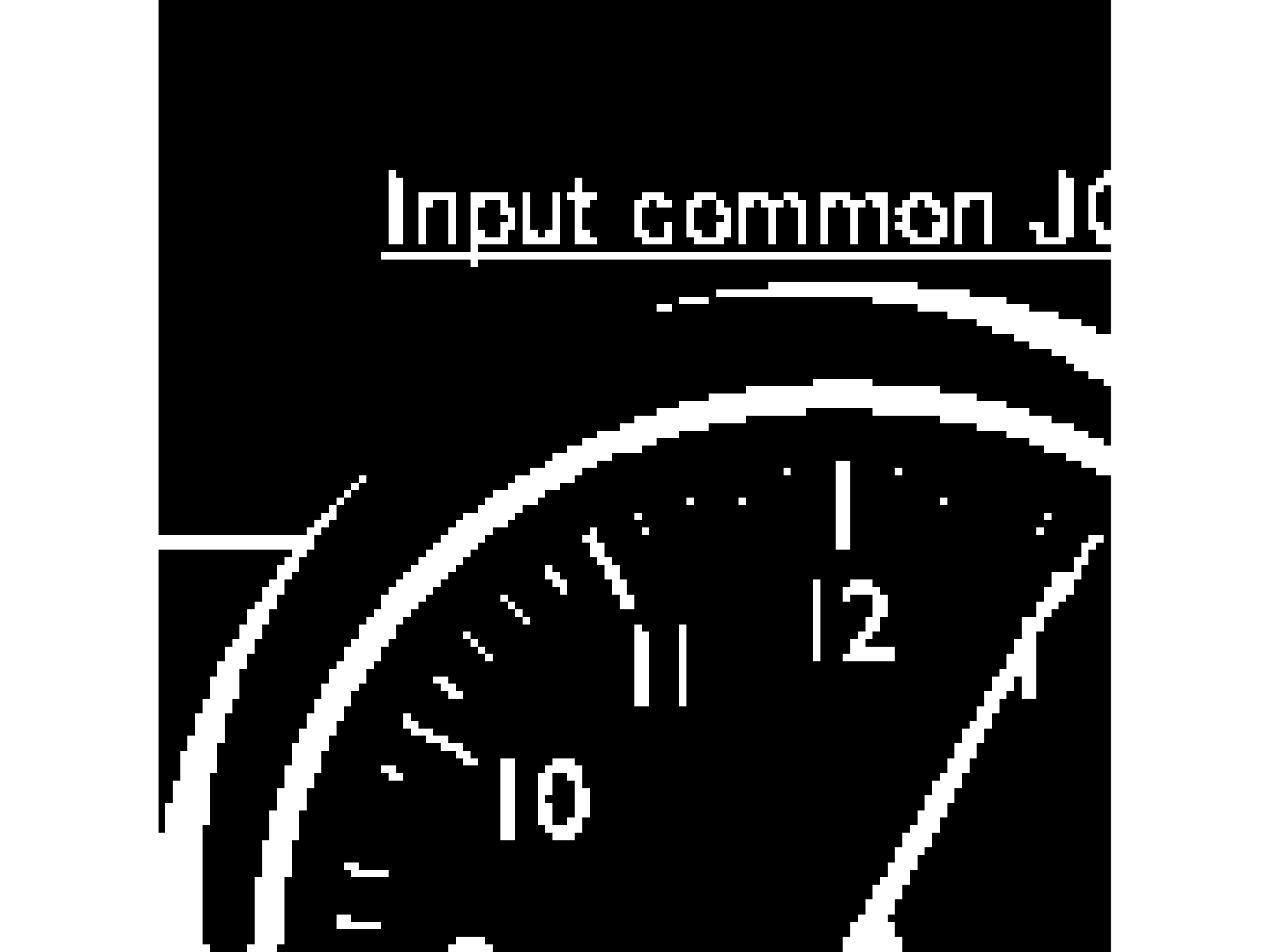}
                 \vspace{-0.45cm}
              \hspace{-4.8cm}
        \end{subfigure} \\[1ex]
        \begin{subfigure}[b]{0.18\textwidth}
                \includegraphics[width=\textwidth]{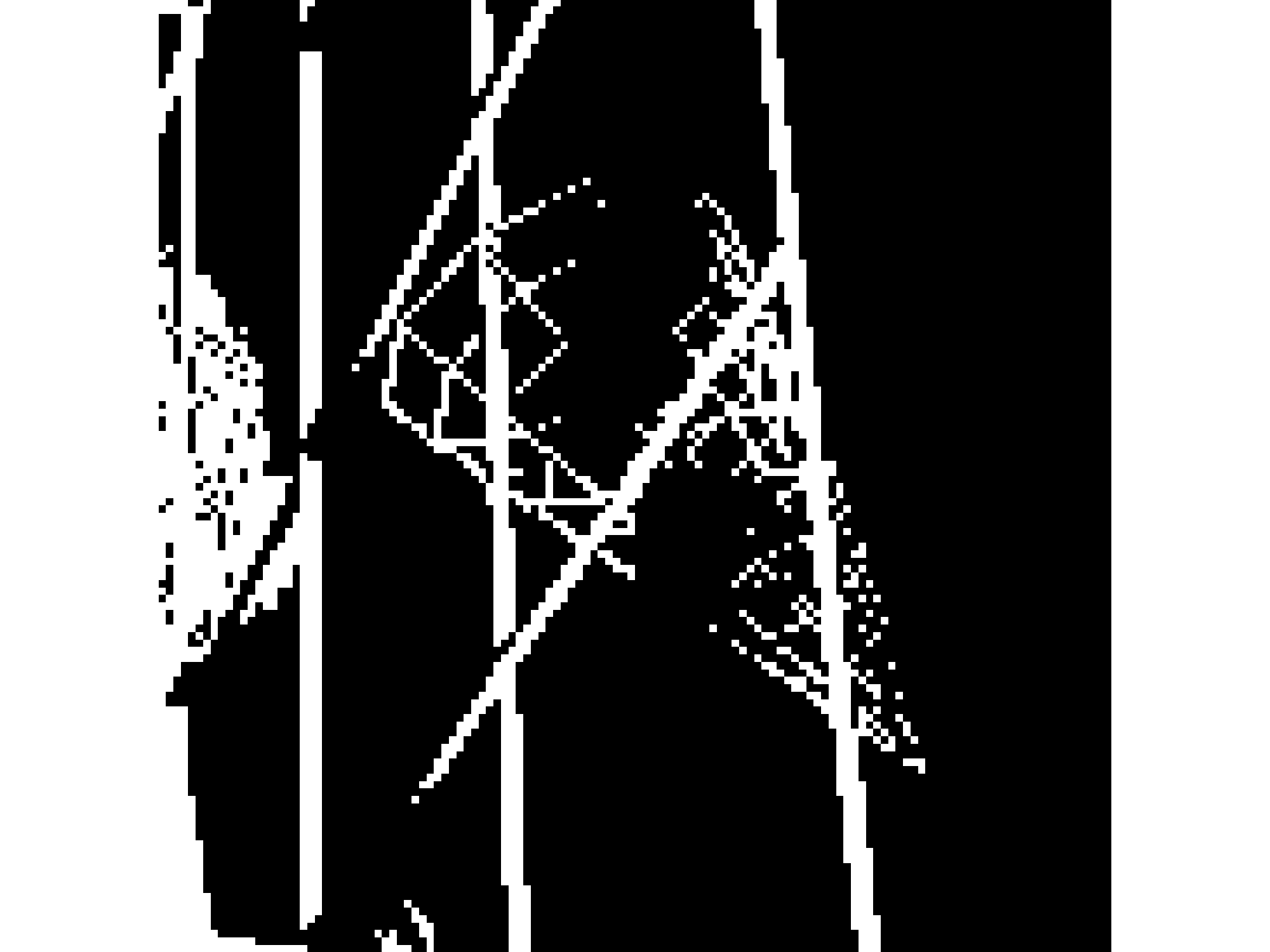}
                                \vspace{-0.5cm}
          \hspace{-2.5cm}    
        \end{subfigure}%
        ~ 
        \begin{subfigure}[b]{0.18\textwidth}
                \includegraphics[width=\textwidth]{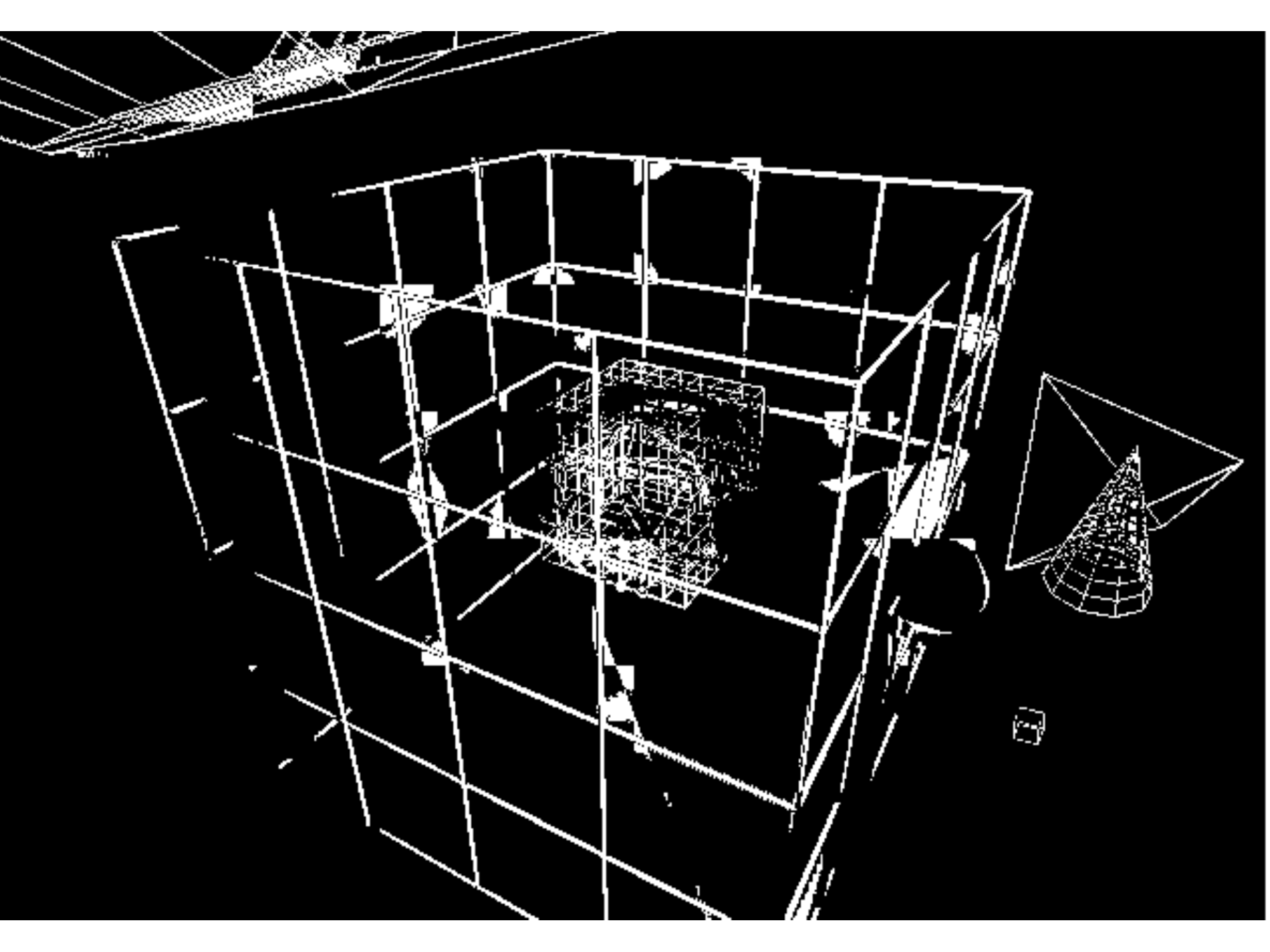}
                \vspace{-0.5cm}
            \hspace{-3cm} 
        \end{subfigure}%
        ~ 
        \begin{subfigure}[b]{0.18\textwidth}
                \includegraphics[width=\textwidth]{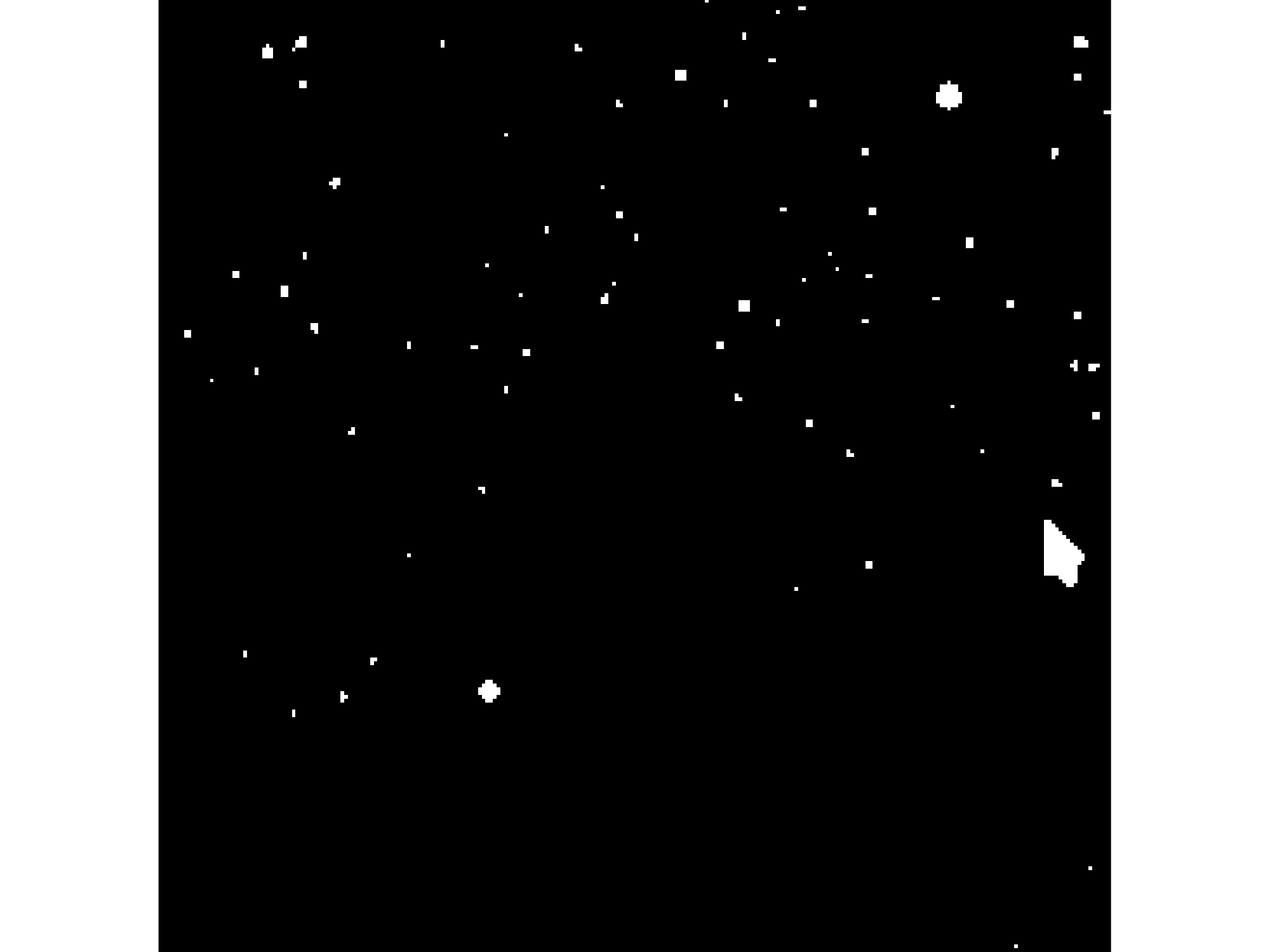}
                \vspace{-0.45cm}
            \hspace{-3cm} 
        \end{subfigure}%
        \begin{subfigure}[b]{0.18\textwidth}
			~ 
                \includegraphics[width=\textwidth]{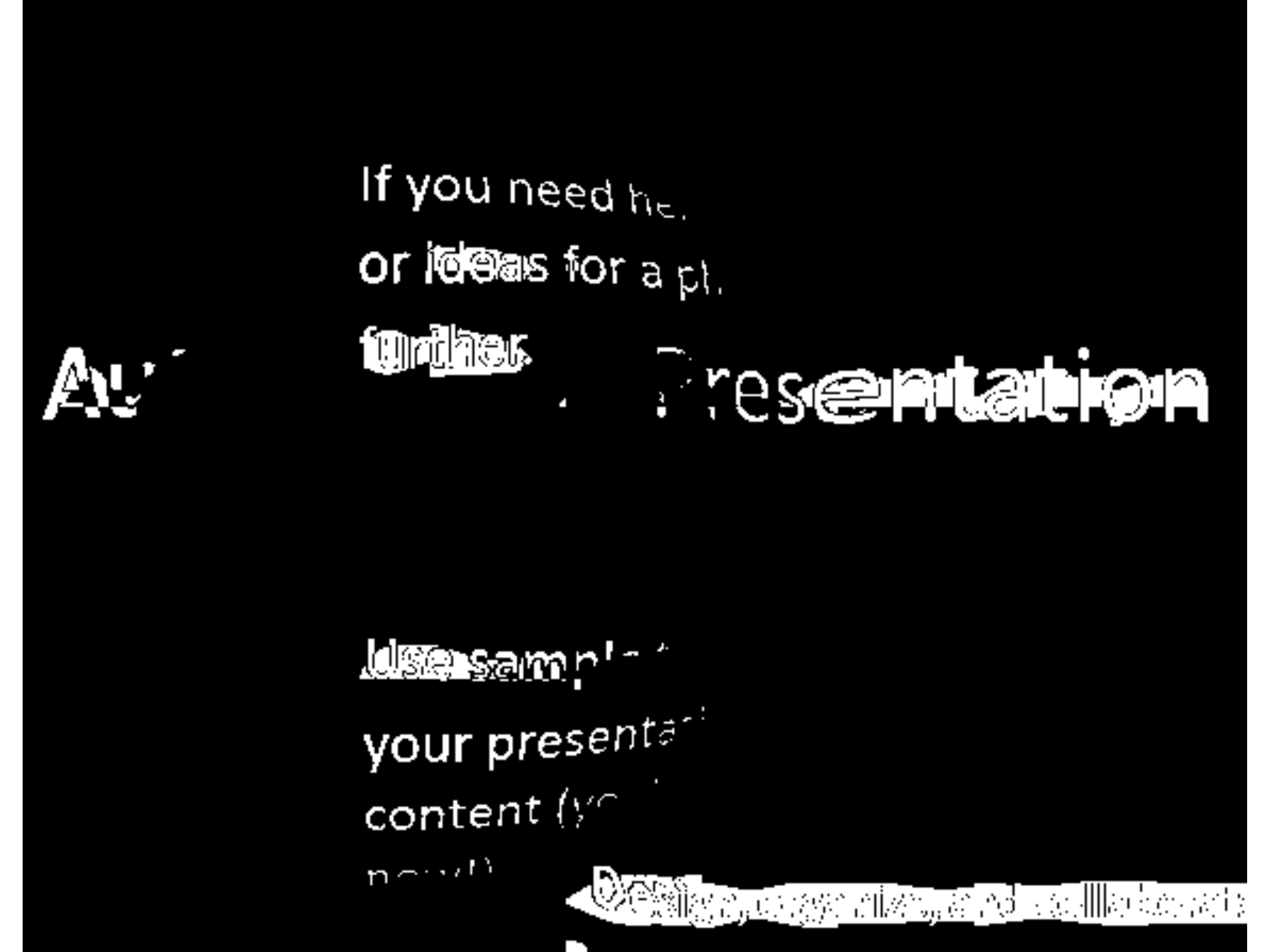}
                \vspace{-0.45cm}
            \hspace{-3cm} 
        \end{subfigure}%
        \begin{subfigure}[b]{0.18\textwidth}
                \includegraphics[width=\textwidth]{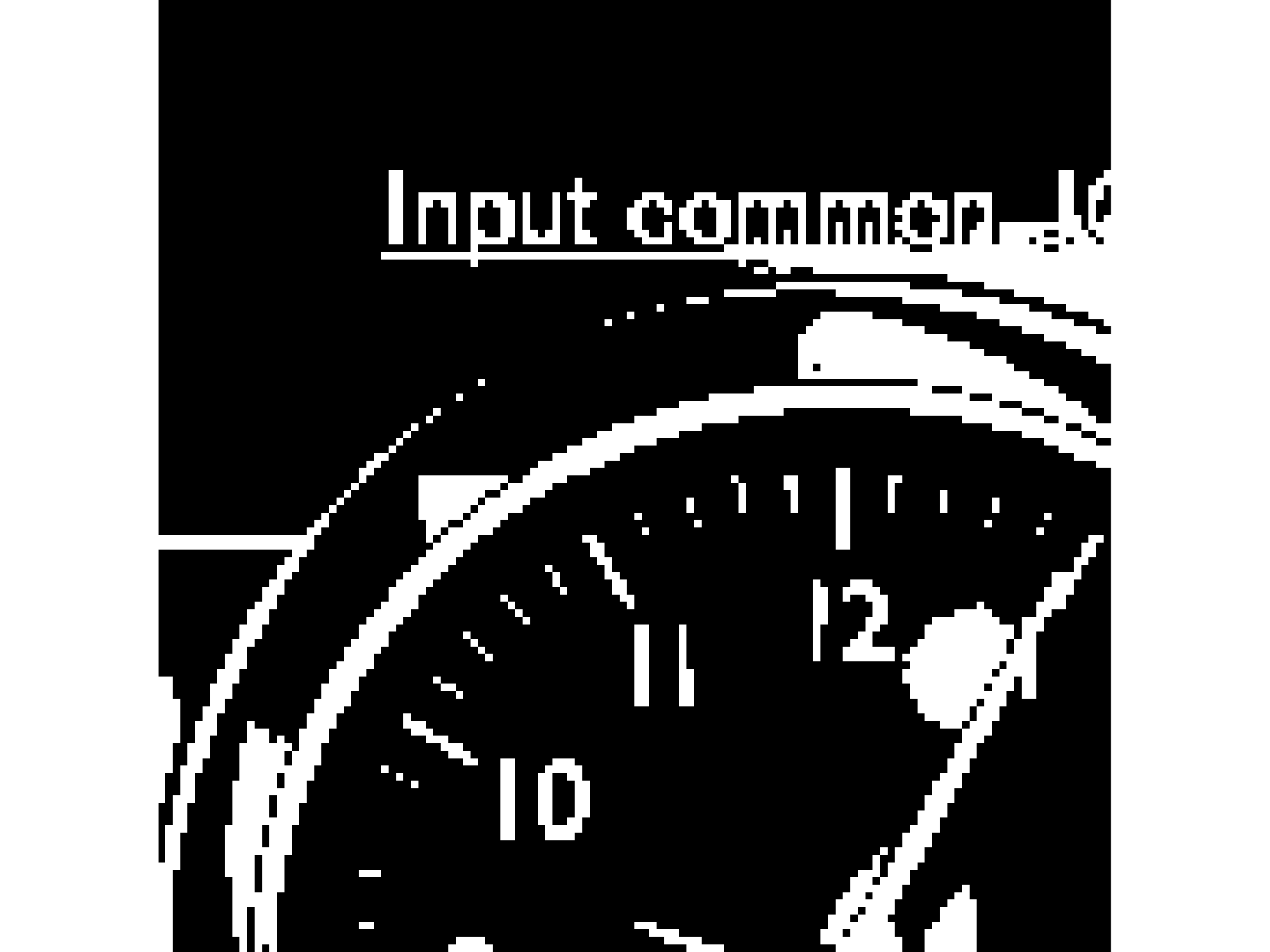}
                 \vspace{-0.45cm}
              \hspace{-4.8cm}
        \end{subfigure} \\[1ex]        
        \begin{subfigure}[b]{0.18\textwidth}
                \includegraphics[width=\textwidth]{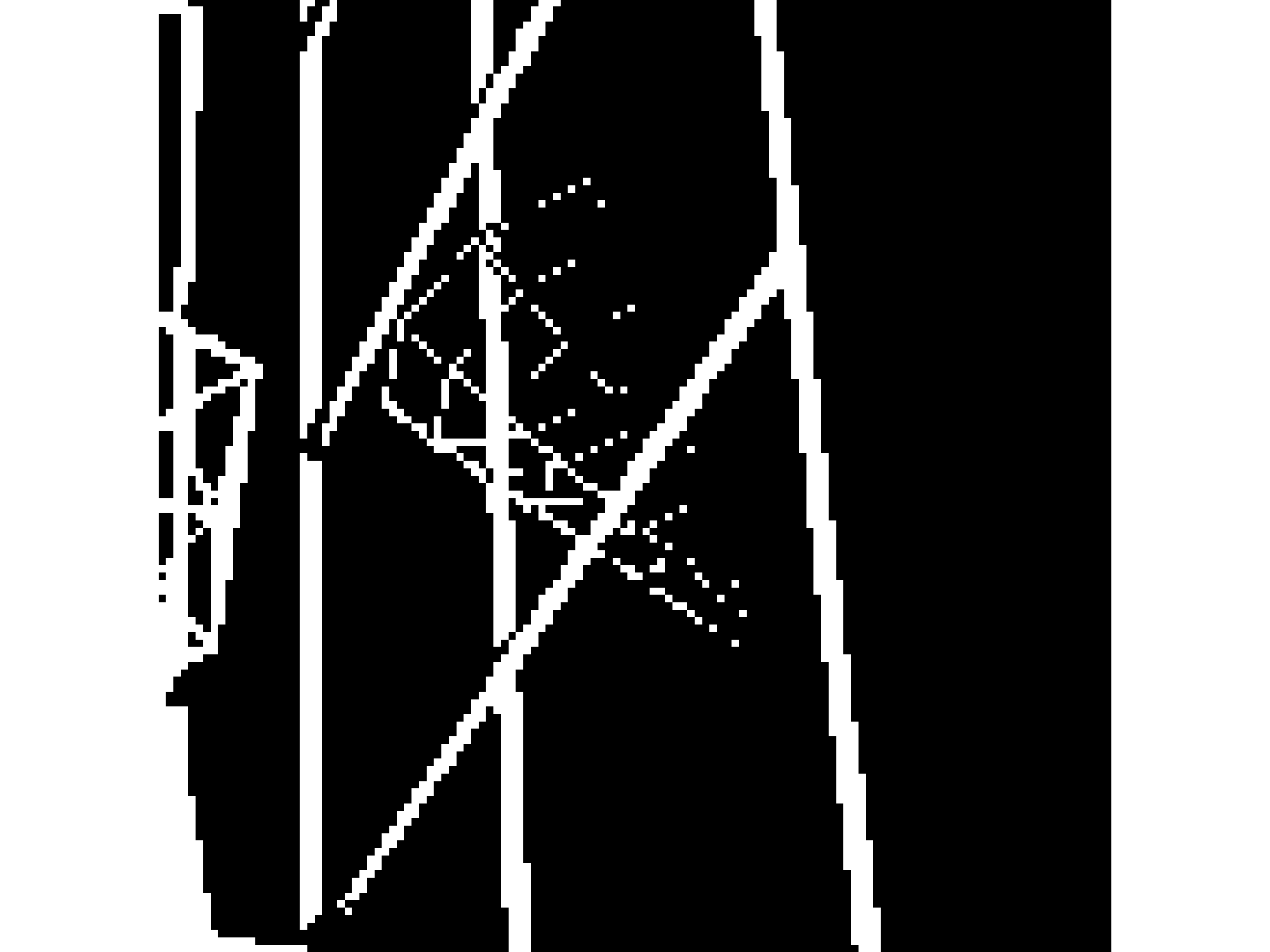}
                                \vspace{-0.5cm}
          \hspace{-2.5cm}    
        \end{subfigure}%
        ~ 
        \begin{subfigure}[b]{0.18\textwidth}
                \includegraphics[width=\textwidth]{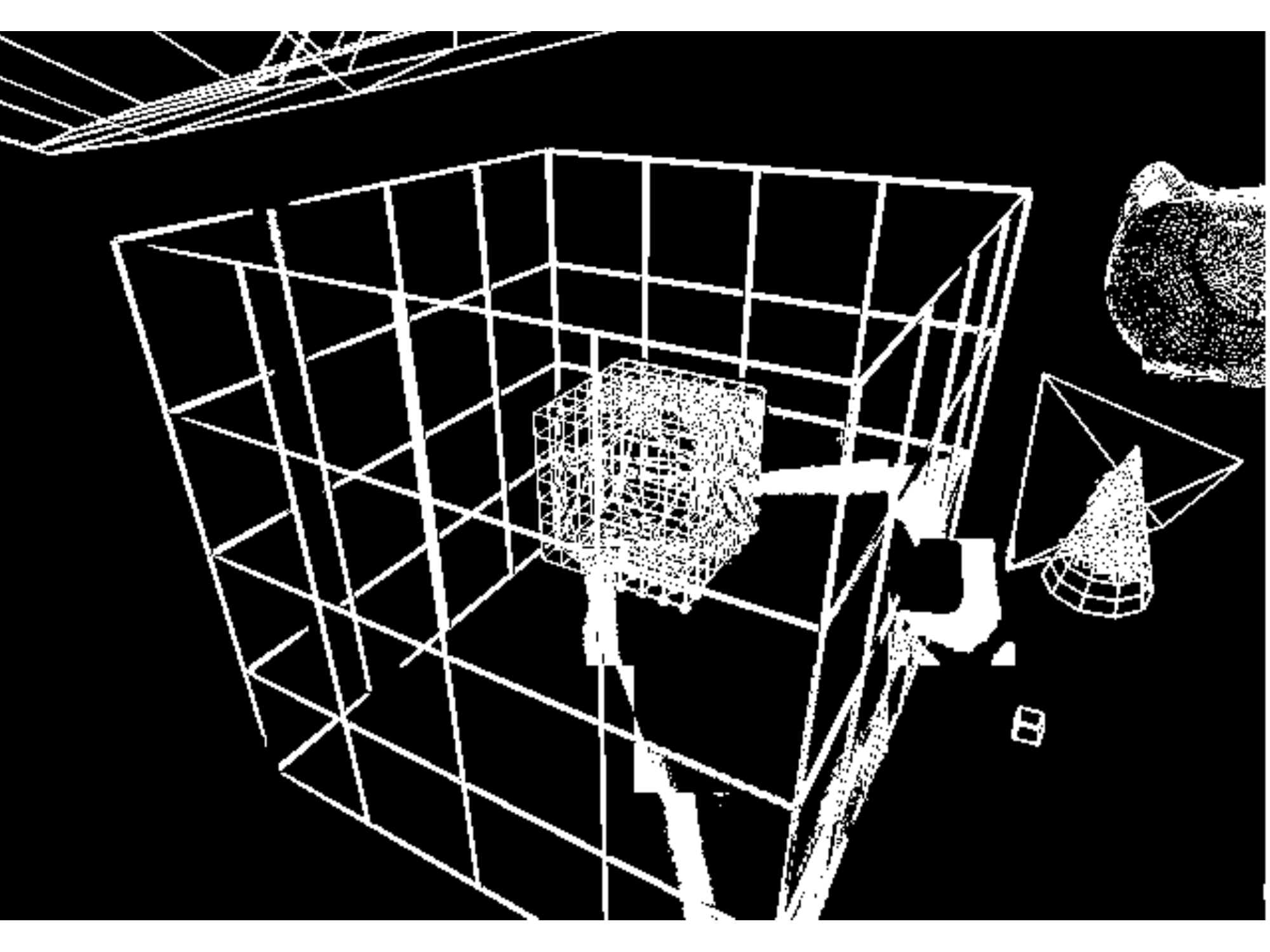}
                \vspace{-0.5cm}
            \hspace{-3cm} 
        \end{subfigure}%
        ~ 
        \begin{subfigure}[b]{0.18\textwidth}
                \includegraphics[width=\textwidth]{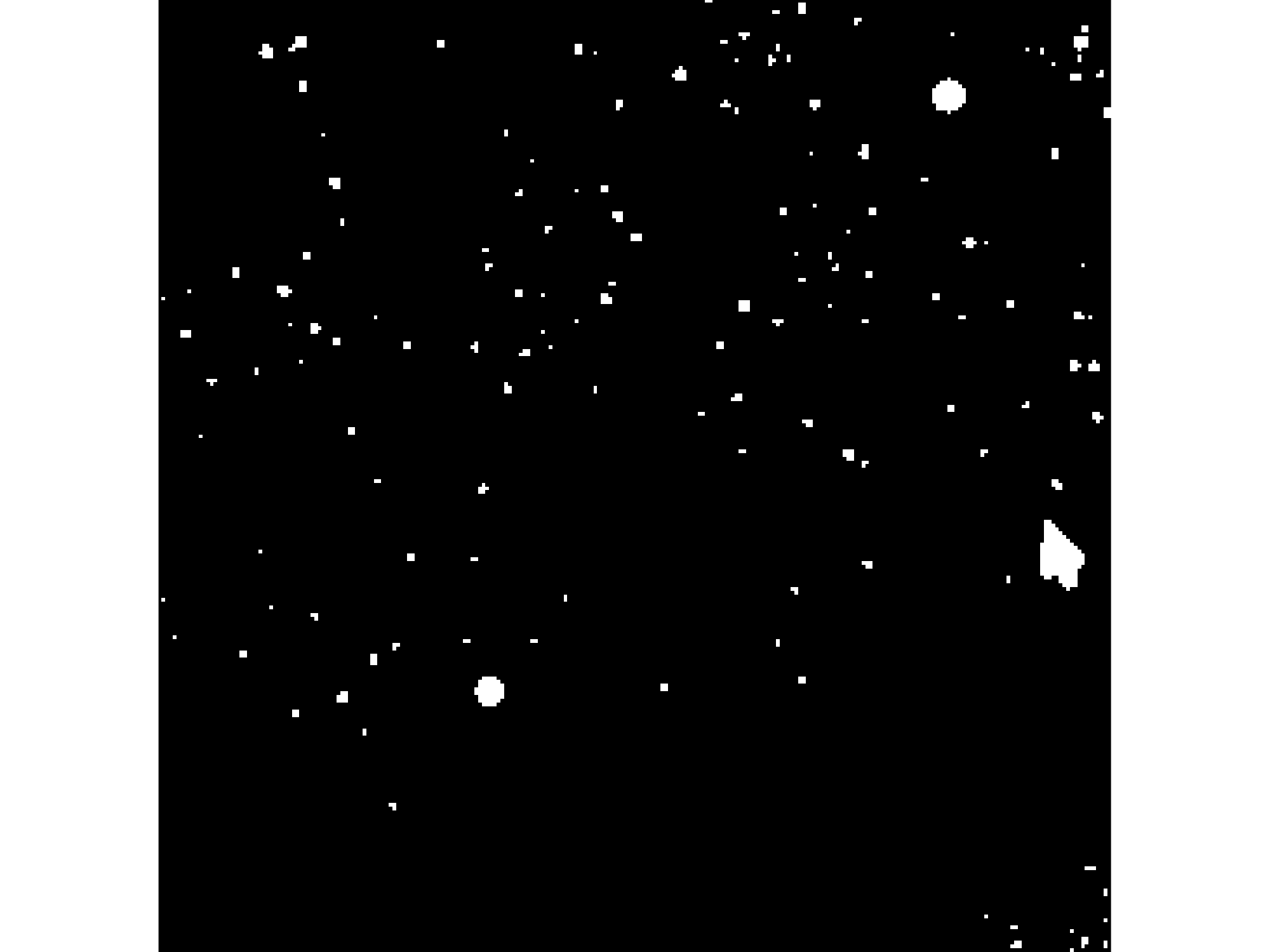}
                \vspace{-0.45cm}
            \hspace{-3cm} 
        \end{subfigure}%
        \begin{subfigure}[b]{0.18\textwidth}
			~ 
                \includegraphics[width=\textwidth]{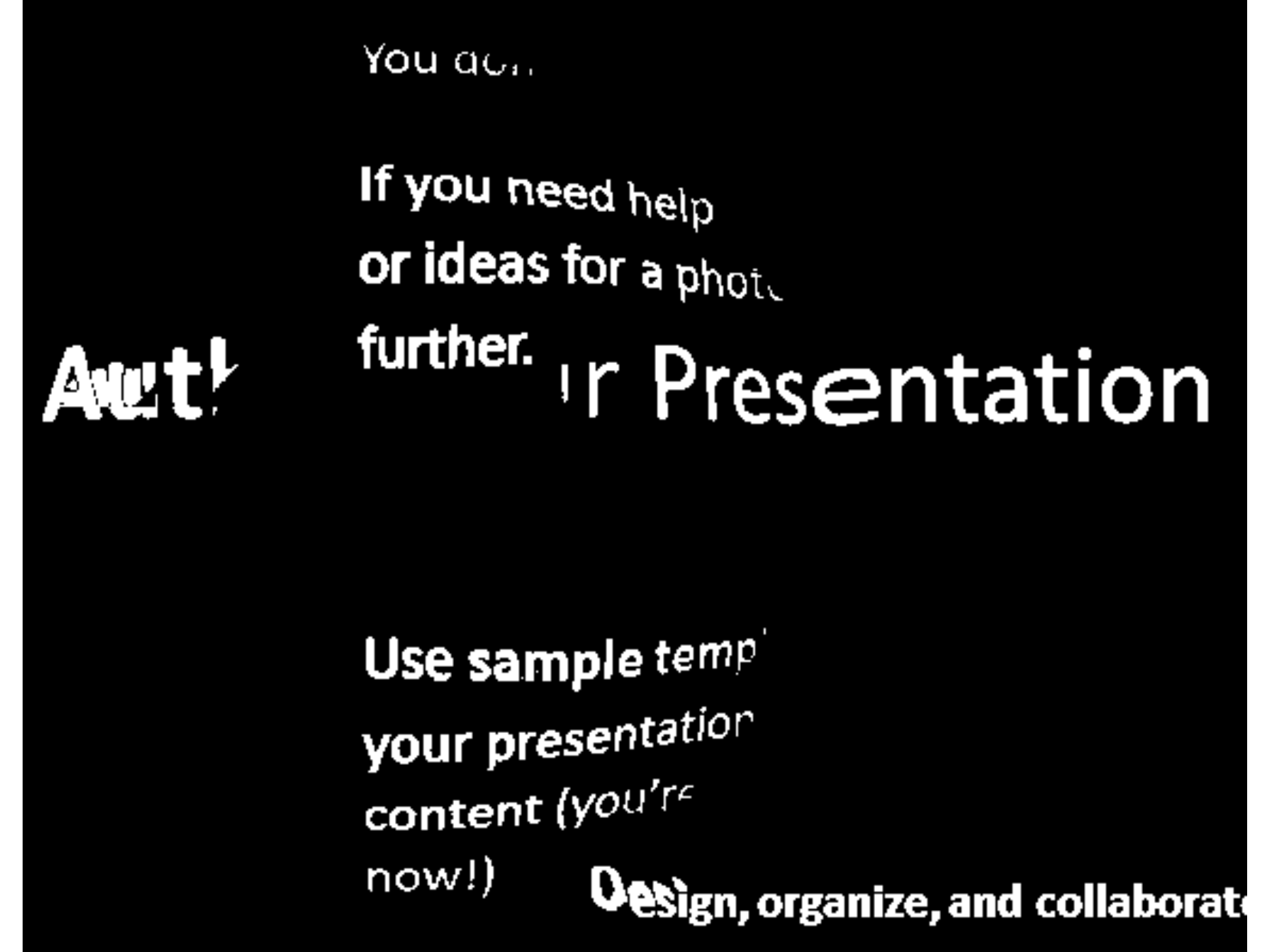}
                \vspace{-0.45cm}
            \hspace{-3cm} 
        \end{subfigure}%
        \begin{subfigure}[b]{0.18\textwidth}
                \includegraphics[width=\textwidth]{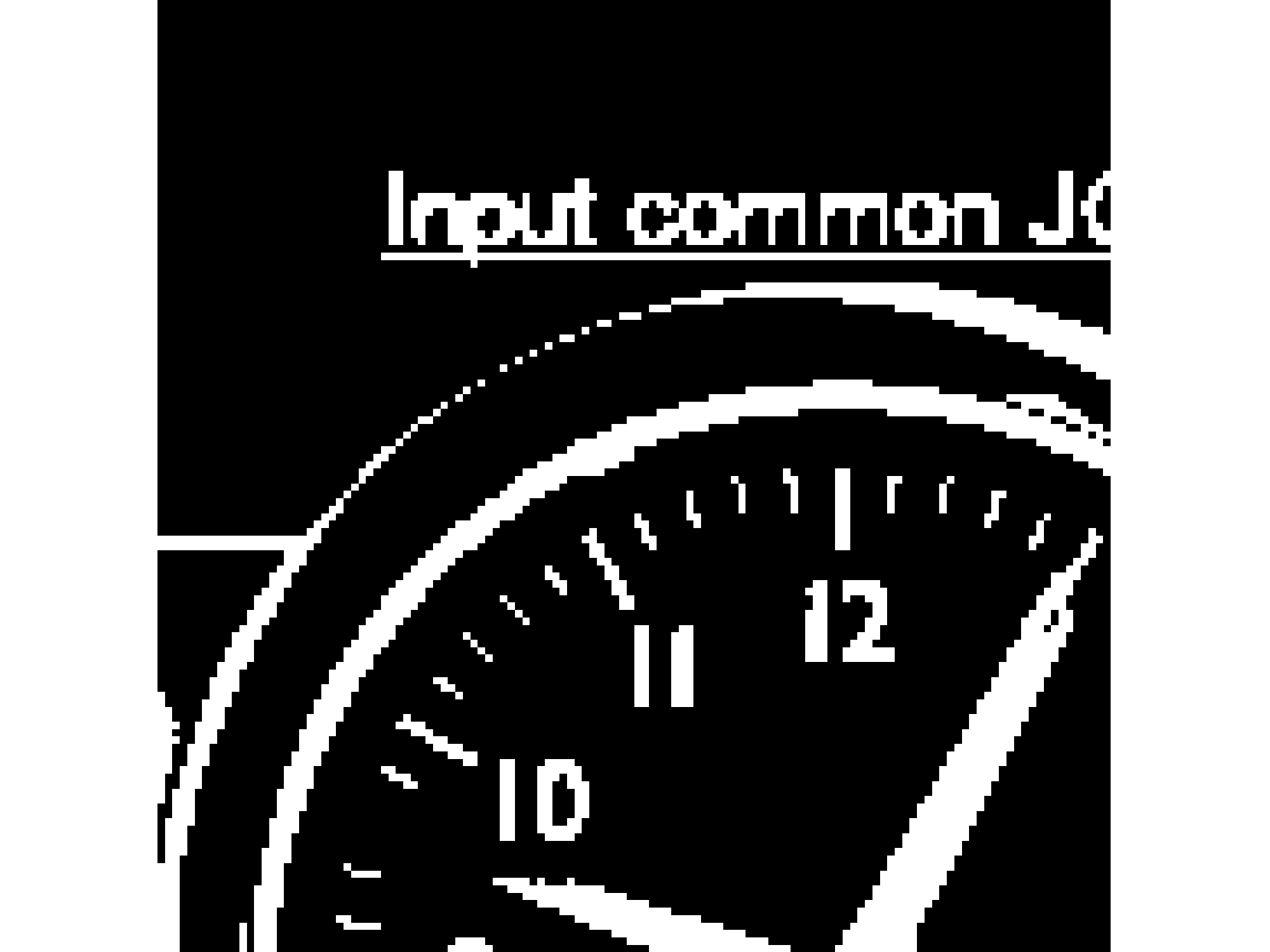}
                 \vspace{-0.45cm}
              \hspace{-4.8cm}
        \end{subfigure} \\[1ex]        
        \begin{subfigure}[b]{0.18\textwidth}
                \includegraphics[width=\textwidth]{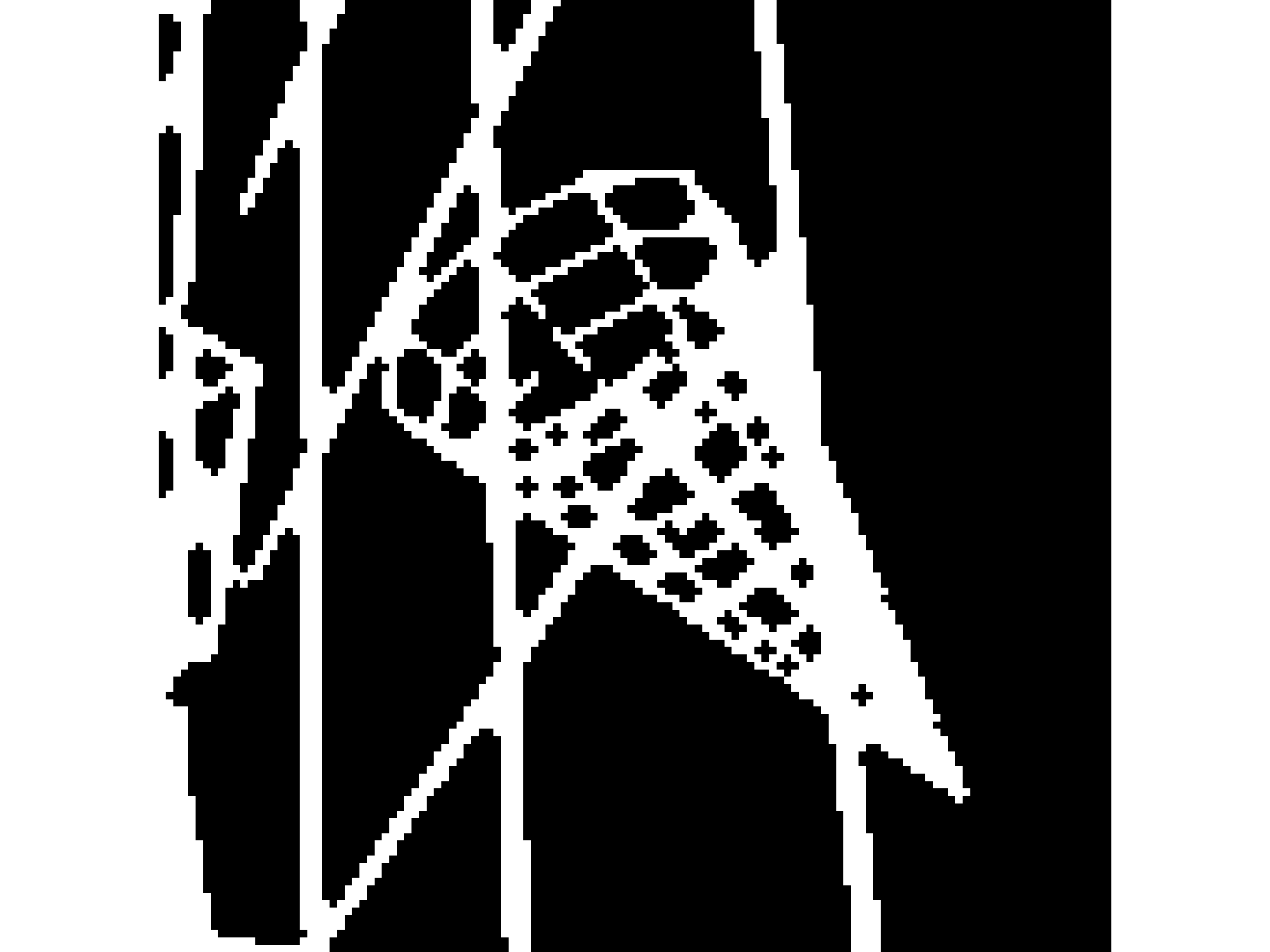}
                                \vspace{-0.5cm}
          \hspace{-2.5cm}    
        \end{subfigure}%
        ~ 
        \begin{subfigure}[b]{0.18\textwidth}
                \includegraphics[width=\textwidth]{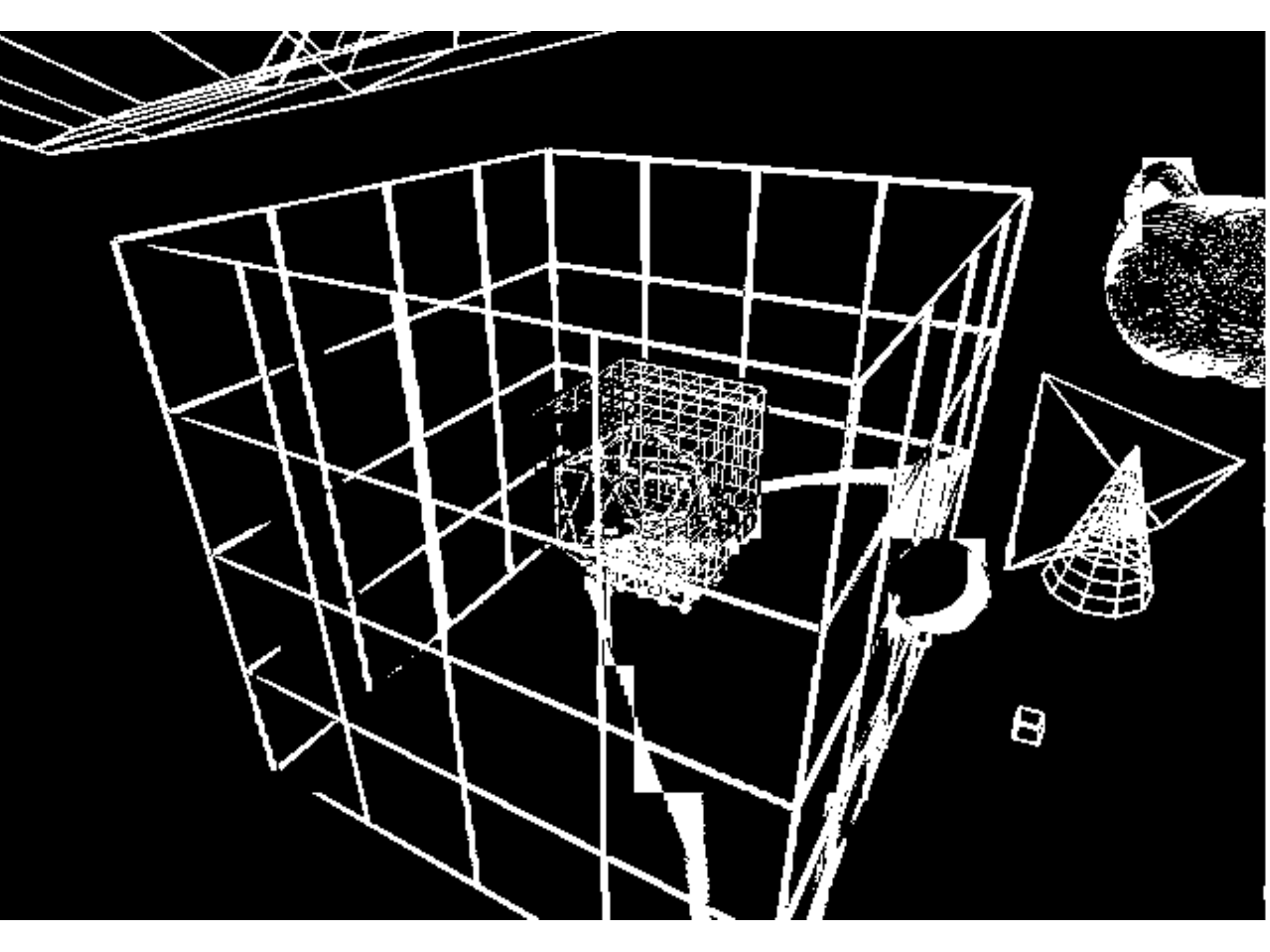}
                \vspace{-0.5cm}
            \hspace{-3cm} 
        \end{subfigure}%
        ~ 
        \begin{subfigure}[b]{0.18\textwidth}
                \includegraphics[width=\textwidth]{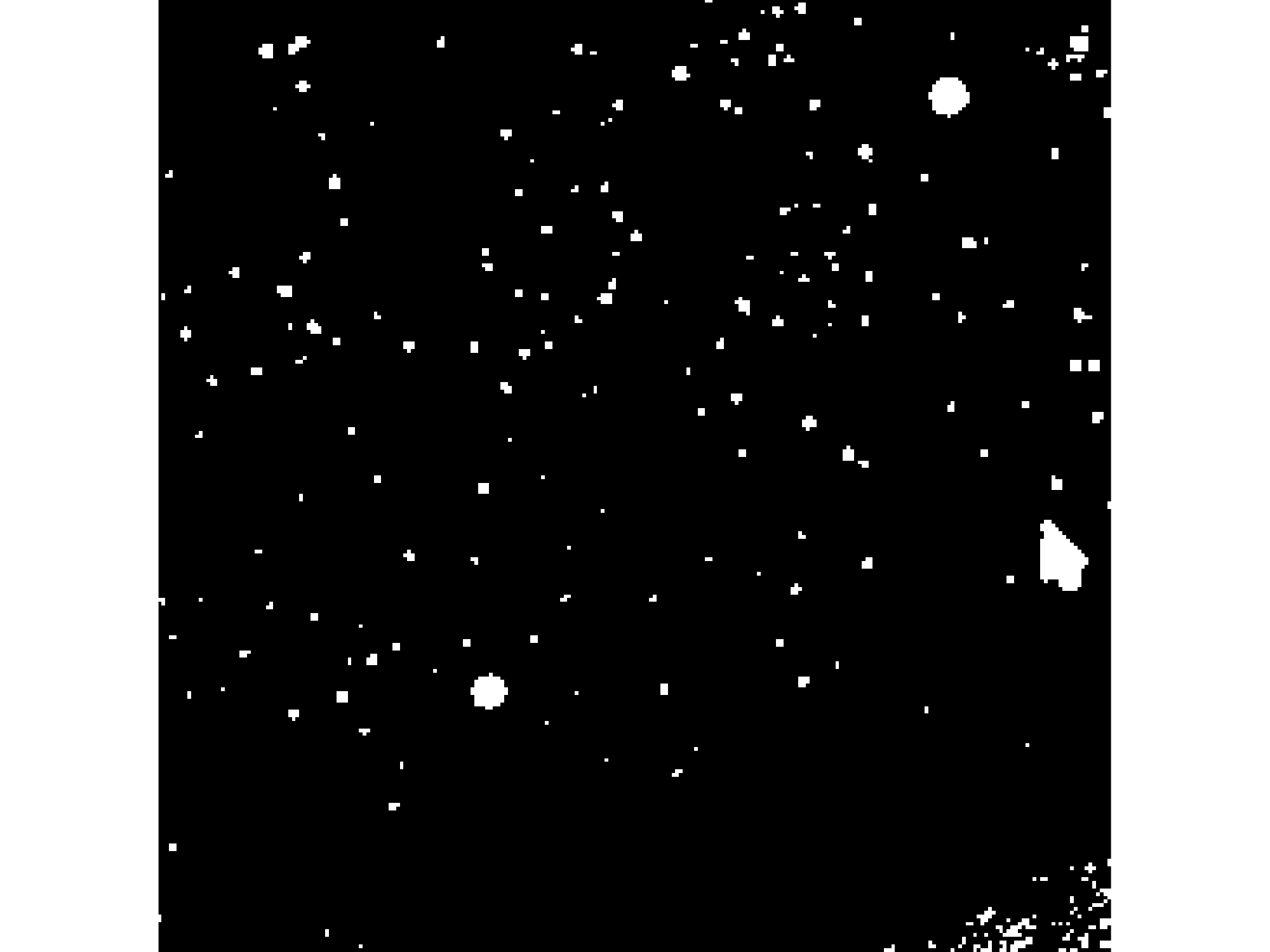}
                \vspace{-0.45cm}
            \hspace{-3cm} 
        \end{subfigure}%
        \begin{subfigure}[b]{0.18\textwidth}
			~ 
                \includegraphics[width=\textwidth]{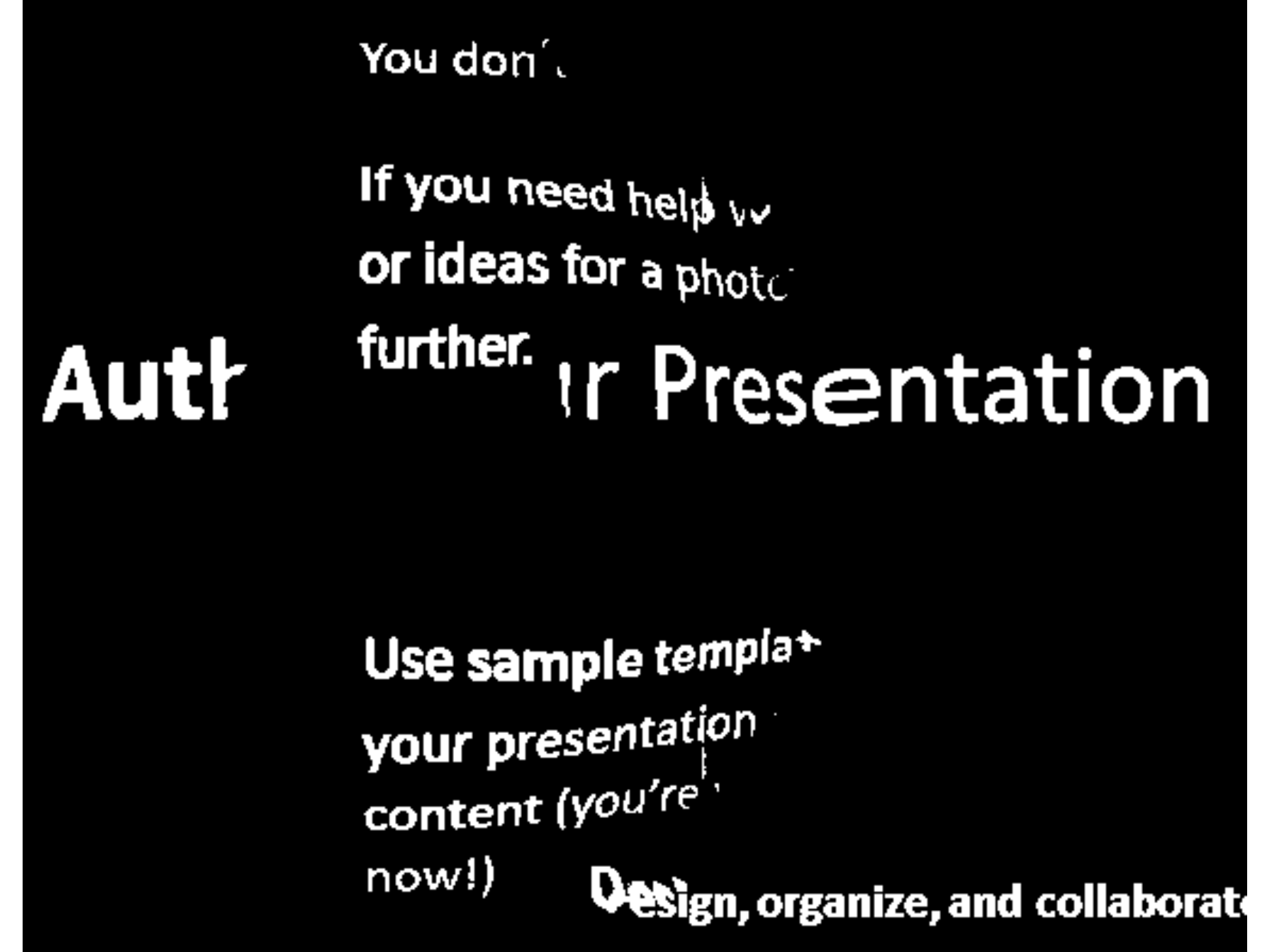} 
                \vspace{-0.45cm}
            \hspace{-3cm} 
        \end{subfigure}%
        \begin{subfigure}[b]{0.18\textwidth}
                \includegraphics[width=\textwidth]{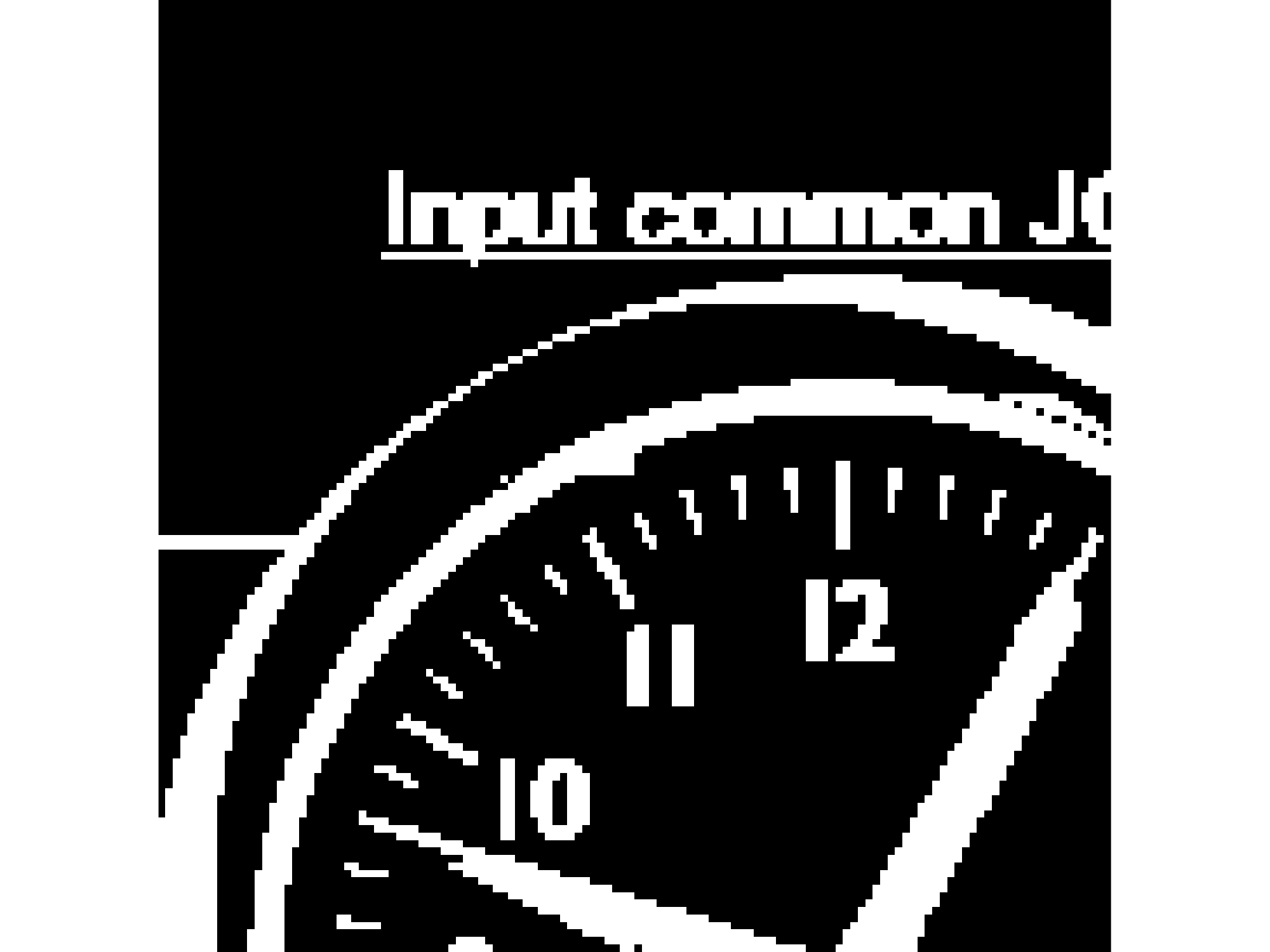}
                 \vspace{-0.45cm}
              \hspace{-4.8cm}
        \end{subfigure}
        \caption{Segmentation result for the selected test images. The images in the first row denotes the original images. The images in the second, third, forth and the fifth rows denote the foreground map by shape primitive extraction and coding, hierarchical clustering in DjVu, least square fitting, and least absolute deviation fitting approach respectively. The images in the sixth row denotes the result by the proposed algorithm.}
\end{figure*}

It can be seen that in all cases the proposed algorithm gives superior performance over DjVu and SPEC. 
Note that our dataset mainly consists of challenging images where the background and foreground have overlapping color ranges. For simpler cases where the background has a narrow color range that is quite different from the foreground, DjVu and least absolute deviation fitting will also work well. 
On the other hand, SPEC usually has problem for the cases where the foreground text/lines have varying colors and are overlaid on  a smoothly varying background.

\section{Conclusion}
This paper proposed an image decomposition scheme that segments an image into background and foreground layers. The background is defined as the smooth component of the image that can be well modeled by a set of DCT functions and foreground as those pixels that cannot be modeled with this smooth representation. We propose to use a robust regression algorithm to fit a set of smooth functions to the image and detect the outliers. The outliers are considered as the foreground pixels. RANSAC algorithm is used to solve this problem.
Instead of applying these robust regression algorithms to every block, which are computationally demanding, we first check whether the block satisfy several conditions and can be segmented using simple methods.

\section*{Acknowledgment}
The authors would like to thank JCT-VC group for providing the HEVC test sequences for screen content coding.
We would also like to thank Huawei Technologies Co., for supporting this work.

\ifCLASSOPTIONcaptionsoff
  \newpage
\fi



\end{document}